\documentclass[10pt,twocolumn,letterpaper]{article}

\usepackage{iccv}
\usepackage{times}
\usepackage{epsfig}
\usepackage{graphicx}
\usepackage{amsmath}
\usepackage{amssymb}

\usepackage[accsupp]{axessibility}
\usepackage{multirow}
\usepackage{array}
\usepackage{float}
\usepackage{bbm}
\usepackage{graphics}
\usepackage{booktabs}
\usepackage{algorithm}
\usepackage{algpseudocode}
\newcommand{\drule}{\specialrule{0.2pt}{1pt}{1pt} \specialrule{0.2pt}{0pt}{\belowrulesep}}
\usepackage{pifont}
\usepackage{xcolor}
\usepackage{pdfrender}

\usepackage{colortbl}

\newcommand*{\affmark}[1][*]{\textsuperscript{#1}}

\usepackage[pagebackref=true,breaklinks=true,letterpaper=true,colorlinks,bookmarks=false]{hyperref}

\iccvfinalcopy 

\ificcvfinal\fi

\begin{document}

\title{CAFA: Class-Aware Feature Alignment for Test-Time Adaptation}

\author{
Sanghun Jung\affmark[1]  \ Jungsoo Lee\affmark[2] \ Nanhee Kim\affmark[3] \ Amirreza Shaban\affmark[1] \ Byron Boots\affmark[1] \ Jaegul Choo\affmark[2] \\
\affmark[1] University of Washington, \ \affmark[2]KAIST AI, \ \affmark[3]Enssel Inc.\\
}

\maketitle
\ificcvfinal\thispagestyle{empty}\fi

\begin{abstract}
\vspace{-0.2cm}
Despite recent advancements in deep learning, deep neural networks continue to suffer from performance degradation when applied to new data that differs from training data.
Test-time adaptation (TTA) aims to address this challenge by adapting a model to unlabeled data at test time. 
TTA can be applied to pretrained networks without modifying their training procedures, enabling them to utilize a well-formed source distribution for adaptation.
One possible approach is to align the representation space of test samples to the source distribution (\textit{i.e.,} feature alignment).
However, performing feature alignment in TTA is especially challenging in that access to labeled source data is restricted during adaptation. 
That is, a model does not have a chance to learn test data in a class-discriminative manner, which was feasible in other adaptation tasks (\textit{e.g.,} unsupervised domain adaptation) via supervised losses on the source data.
Based on this observation, we propose a simple yet effective feature alignment loss, termed as Class-Aware Feature Alignment (CAFA), which simultaneously 1) encourages a model to learn target representations in a class-discriminative manner and 2) effectively mitigates the distribution shifts at test time.
Our method does not require any hyper-parameters or additional losses, which are required in previous approaches.
We conduct extensive experiments on 6 different datasets and show our proposed method consistently outperforms existing baselines.
\end{abstract}

\vspace{-0.5cm}
\section{Introduction}
\vspace{-0.2cm}
\label{sec:introduction}
Recent advancements~\cite{resnet, transformer, vit, bert} in machine learning are effective in solving diverse problems, achieving remarkable performance enhancements on benchmark datasets.
However, these methods can suffer from significant performance degradation when applied to test data with different properties from the training data (\textit{i.e.,} source data), such as corruption~\cite{corruption}, changing lighting conditions~\cite{dist_shift_fail2_light}, or adverse weather~\cite{dist_shift_fail3_weather, robustnet}.
Sensitivity to distribution shifts~\cite{datasetshift} hampers deep networks from performing well in practical scenarios where test samples may differ from training data~\cite{dist_shift_fail1}.
Thus, adapting deep models to the test samples is crucial when distribution shifts exist.

Various adaptation methods~\cite{uda_bousmalis, uda_gong, uda_long, dann, coral, uda_fa_ganin, sfda_xia, sfda_historical} have been proposed to alleviate this problem.
However, most of these methods require either access to the source data during adaptation~\cite{coral, dann, cycada} or modification of the training procedure~\cite{ttt++, shot, ttt}, which limits their applicability.
Therefore, we seek to design an adaptation method that 1) is applicable to existing deep networks without modification and 2) does not require access to the source data during adaptation.
Satisfying such conditions, previous studies perform adaptation at test time while making predictions simultaneously, which is referred to as \emph{test-time adaptation} (TTA).

One widely adopted approach to address distribution shifts is to align the source (\textit{i.e.,} training data) and target (\textit{i.e.,} test data) distributions~\cite{dann, coral, uda_fa_long, uda_fa_dirt, uda_fa_ganin, uda_fa_tzeng}.
For example, DANN~\cite{dann} directly reduces the $\mathcal{H}$-divergence between the source and target distributions, and CORAL~\cite{coral} minimizes the difference in the second-order statistics between the source and target data.
Despite their demonstrated effectiveness in the unsupervised domain adaptation (UDA) task, applying those alignments to TTA has the following limitation.
Alignments are generally performed along with supervised losses on the source data, which encourages a model to learn target distributions in a class-discriminative manner~\cite{coral}.
However, access to the source data is prohibited during adaptation in TTA, precluding learning class discriminability.

\begin{figure*}[t!]
\centering
  \includegraphics[width=0.9\linewidth]{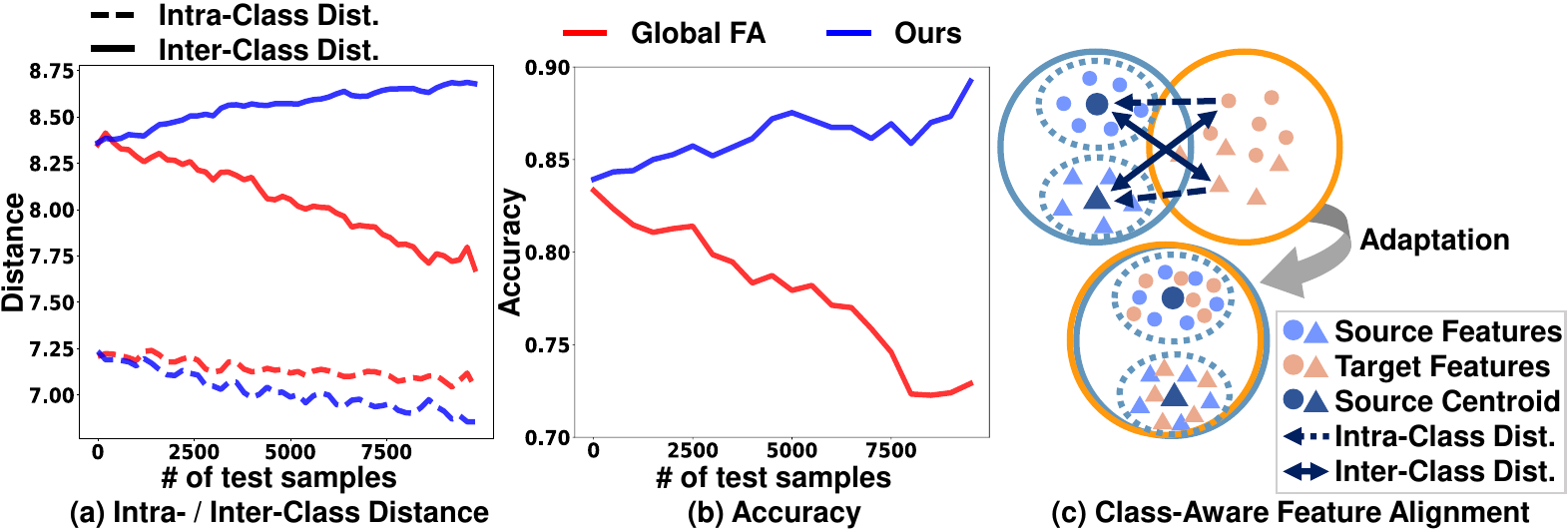}
    \caption{A motivating example of our paper. (a) shows the change of the intra-class distance (dotted lines) and the inter-class distance (solid lines) of ours (blue line) and global feature alignment (red line). (b) shows the accuracy changes as adaptation proceeds.
    (c) illustrates how our method CAFA aligns the test features to the source distribution in a class-discriminative manner.
    We obtain the plots by adapting a model to corrupted images of the CIFAR10-C dataset.
    Please refer to Section~\ref{sec:experiments} for further details.
    }
    \vspace{-0.3cm}
    \label{fig:motivation}
\end{figure*}

With this issue in mind, we conduct an analysis of the effects of feature alignments in TTA by using two distances in the representation space: intra-class distance and inter-class distance.
As shown in Fig.~\ref{fig:motivation} (c), intra-class distance (dotted arrow) is defined as the distance between a sample and its ground-truth source class distribution, and inter-class distance (solid arrow) denotes the averaged distance between the sample and the other source class distributions.
Achieving low intra-class distance and high inter-class distance is crucial for improving classification accuracy~\cite{lee-face, uda_cfa_chen, uda_cfa_li, uda_cfa_pl1, uda_cfa_xie}.

For analysis, we first adopt a feature alignment that reduces the domain-level discrepancy between source and target domains, which is a commonly adopted paradigm in UDA studies~\cite{dann, coral, uda_fa_dirt}.
One straightforward approach to achieve this is to align the mean and covariance of the source and target distributions, \textit{i.e.,} global feature alignment (Global FA).\footnote{We will explain the global feature alignment in more detail in Section~\ref{sec:method_analysis}\vspace{-0.9cm}}
The result of applying this feature alignment in TTA is depicted in Fig.~\ref{fig:motivation} (a) (red line).
The intra-class distance is reduced, which is desirable but is also accompanied by a decrease in inter-class distance.
Such effects can degrade the image classification accuracy in Fig.~\ref{fig:motivation} (b) (red line).
This is mainly due to the lack of class information in the global feature alignment.
A model does not have a chance to learn the test data in a class-discriminative manner since a supervised loss is not available on both the source and target data.

Motivated by such observations, we propose Class-Aware Feature Alignment (CAFA) that aligns the target features to the pre-calculated source feature distributions by considering both intra- and inter-class distances.
To be more specific, we pre-calculate the statistics (\textit{i.e.,} mean and covariance) of the source distribution to estimate class-conditional Gaussian distributions from a pretrained network.
At test time, we use the Mahalanobis distance~\cite{mahalanobis} to 1) align each sample to its predicted class-conditional Gaussian distribution (\textit{i.e.,} reduce intra-class distance) and 2) enforce samples to be distinct from the other class-conditional Gaussian distributions (\textit{i.e.,} increase inter-class distance).
Applying CAFA successfully enhances class discriminability as shown in Fig.~\ref{fig:motivation} (a) (blue line) and significantly improves the classification accuracy as adaptation proceeds (Fig.~\ref{fig:motivation} (b) (blue line)). We empirically show that reducing intra-class distance alone is not sufficient as it could also reduce the inter-class distance and result in performance degradation.

Aligning feature distributions at test time requires access to the source data \textit{before adaptation} to pre-calculate the source statistics, as similarly done in previous methods~\cite{bufr, ttt++, sungha_eccv}.
However, we empirically show that CAFA only requires a small number of training samples (\textit{e.g.,} 5\% of the training samples in the ImageNet/CIFAR10 datasets (Fig.~\ref{fig:train_sample})) to obtain robust source statistics that outperform the existing methods.
In addition, CAFA does not require any hyper-parameters or modifications on pretraining procedures for adaptation.

The main contributions of our work are as follows:
\begin{itemize}
    \vspace{-0.2cm}
    \item We propose a novel Class-Aware Feature Alignment (CAFA) that effectively mitigates distribution shifts and encourages a model to learn discriminative target representations simultaneously.
    \vspace{-0.2cm}
    \item Our proposed approach is simple yet effective, not requiring hyper-parameters or additional modifications of the training procedure.
    \vspace{-0.2cm}
    \item We conduct extensive experiments on 6 different datasets along with in-depth analyses and show that CAFA consistently outperforms the existing methods on test-time adaptation.
    \vspace{-0.2cm}
\end{itemize}

\section{Related Work}
\label{sec:related_work}
\subsection{Test-time Adaptation}
Existing UDA approaches~\cite{ben_uda, ben_uda2, uda_bousmalis, uda_gong, uda_long, uda_pinheiro, uda_saito, cycada} have addressed distribution shifts effectively by adapting to target domains at training time.
UDA approaches generally assume that 1) source data is available during adaptation, and 2) we already know which target domain the models are adapted to.
However, these assumptions sometimes do not hold in real-world scenarios.
To address such concerns, approaches that adapt a model at test time have been proposed, not requiring access to the source data during adaptation~\cite{tent, mixnorm, tta_batch_stat, memo, norm, pl, ttt, ttt++, t3a}.
Several methods~\cite{ttt, ttt++} perform adaptation in an offline manner, predicting test samples after iterating multiple epochs over the entire set of the test samples (\textit{i.e.,} test-time training).
These approaches modify the training procedure to have self-supervised losses (\textit{e.g.,} rotation prediction or contrastive loss) and utilize them as proxy losses for adaptation.
However, as also pointed out in Wang \textit{et al.}~\cite{tent}, it is not guaranteed that optimizing the proxy losses helps in improving the main task since they are not directly related to classifying images into categories.
Addressing such concerns, test-time adaptation (TTA) methods~\cite{tent, mixnorm, tta_batch_stat, memo, norm, adacontrast} have been proposed.
These approaches do not require any modification of the training procedures, allowing the algorithms to be applicable to a given pretrained deep learning network.
TENT~\cite{tent}, a recent seminal work in TTA, proposed to update the modulation parameters in batch normalization~\cite{batch_norm} layers while minimizing the entropy loss, effectively mitigating distribution shifts.

\subsection{Feature Alignment}
Feature alignment is widely adopted in UDA studies to mitigate distribution shifts~\cite{uda_fa_dirt, uda_fa_ganin, uda_fa_long, uda_fa_tzeng}.
However, most of these approaches do not consider categorical information but rather match the source and target distributions globally.
This may harm class discrimination performance since it does not guarantee class-to-class matching between two distributions~\cite{uda_cfa_chen}.
Tacking the problem, various studies have proposed to align distributions in a class-discriminative manner~\cite{uda_cfa_chen, uda_cfa_li, uda_cfa_long, uda_cfa_haeusser, uda_cfa_pl1, uda_cfa_pl2, uda_cfa_pl3}.
This point of view is also relevant to test-time adaptation, and we design an effective loss that simultaneously mitigates the distribution gap while improving class discriminability.

\begin{figure*}[t!]
\centering
  \includegraphics[width=0.9\linewidth]{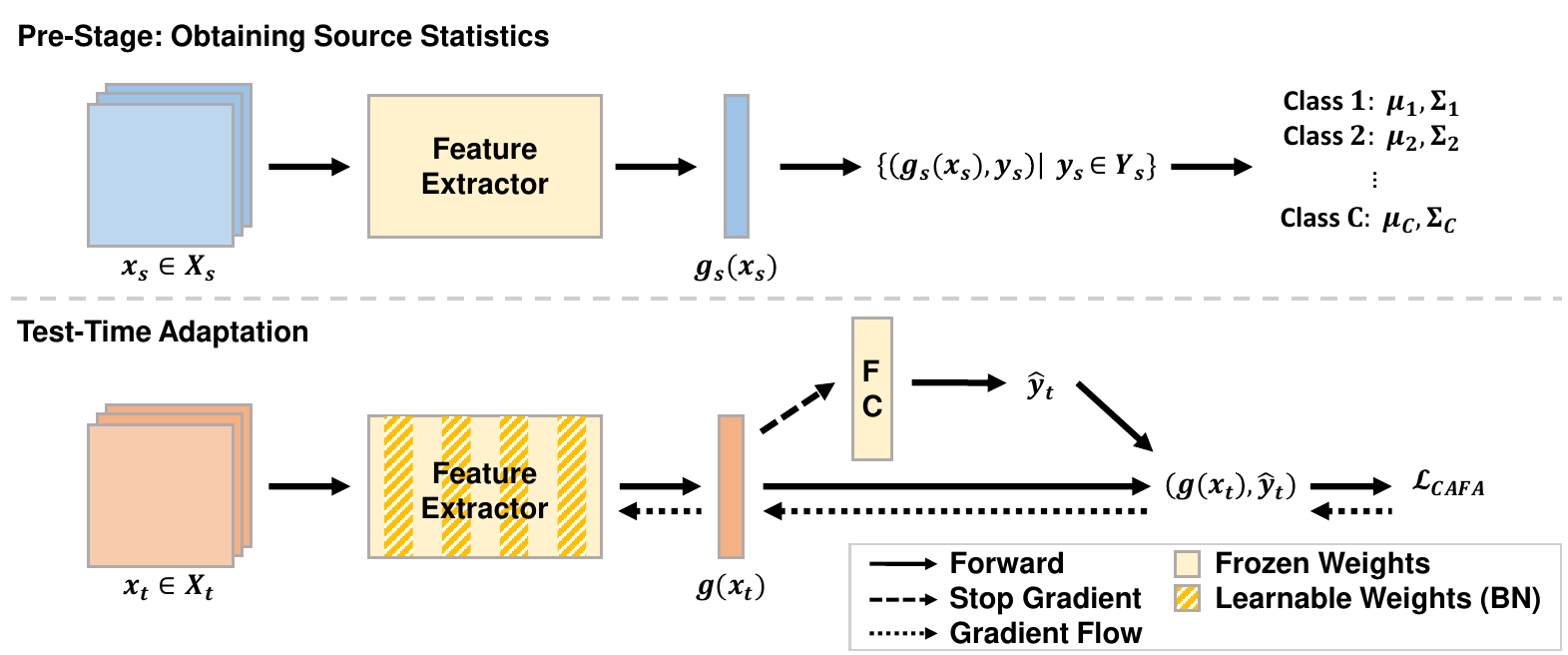}
    \caption{Overview of our method. (Pre-stage) Our method first pre-calculates source class-conditional Gaussian distributions using a pre-trained network. (Test-time adaptation) During test-time adaptation, we adapt a model by optimizing class-aware feature alignment loss while making predictions simultaneously.}
    \vspace{-0.4cm}
    \label{fig:method_overview}
\end{figure*}

\section{Proposed Method}
\label{sec:method}

\subsection{Preliminary}
\label{sec:preliminary}
Assume that we have a model $f_s(x)=h_s\circ{g_s}(x)$ pretrained with a supervised loss $\mathcal{L}(x_s, y_s)$ on source data, where $x_s \in \mathcal{X}_s$ and $y_s \in \mathcal{Y}_s$.
Here $g_s: \mathcal{X}_s \rightarrow \mathbb{R}^d$ denotes the pretrained feature extractor, and $h_s: \mathbb{R}^d \rightarrow \mathbb{R}^C$ indicates the pretrained classifier, where $d$ is the dimension of extracted features, and $C$ is the number of classes.
Then, we aim to adapt the pretrained model $f_s(\cdot)$ to target data $x_t$ while correctly classifying them at test time.

\noindent\textbf{Mahalanobis distance}
In this work, we adopt the Mahalanobis distance~\cite{mahalanobis} to align the source and target distributions.
The Mahalanobis distance measures the distance between a distribution and a sample.
With an input image $x$, feature extractor $g(\cdot)$, and Gaussian distribution $\mathcal{N}(\boldsymbol{\mu}, \boldsymbol{\Sigma})$, the Mahalanobis distance is defined as
\begin{equation}
    D(x; \boldsymbol{\mu}, \boldsymbol{\Sigma}) = (g(x) - \boldsymbol{\mu})^{\top}\boldsymbol{\Sigma}^{-1}(g(x) - \boldsymbol{\mu}).
\end{equation}

\noindent\textbf{Intra-/inter-class distance}
For analysis, we measure the intra- and inter-class distances between the class-conditional source distributions and target samples.
To be more specific, we define class-conditional Gaussian distributions as $P(g_s(x)|y=c) = \mathcal{N}(g_s(x)|\boldsymbol{\mu}_c, \boldsymbol{\Sigma}_c)$, where $\boldsymbol{\mu}_c, \boldsymbol{\Sigma}_c$ are the mean and covariance of the multivariate Gaussian distribution of class $c \in \{1, ..., C\}$.
Then, with a target image $x_t$, the intra-class distance is defined as
\begin{equation}
    D_{\text{intra}}(x_t, y_t) = D(x_t; \boldsymbol{\mu}_{y_t}, \boldsymbol{\Sigma}_{y_t}),
\end{equation}
where $y_t$ indicates the corresponding ground-truth label of the target image.
Analogously, the inter-class distance is defined as 
\begin{equation}
    D_{\text{inter}}(x_t, y_t) = \frac{1}{C - 1}\sum^{C}_{c=1}\mathbbm{1}(y_t \neq c)D(x_t; \boldsymbol{\mu}_{c}, \boldsymbol{\Sigma}_{c})
\end{equation}
Note that achieving low intra-class distance and high inter-class distance is important for improving image classification accuracy.

\subsection{Analysis of Class Discriminability in Feature Alignment}
\label{sec:method_analysis}
We compare and analyze three different feature alignments with respect to the intra- and inter-class distances.
First, we investigate the global feature alignment (Global FA) that reduces the discrepancy between the source and target distributions without considering class information.
With the given source Gaussian distribution $\mathcal{N}(g_s(x_s)|\boldsymbol{\mu}_s, \boldsymbol{\Sigma}_s)$, the Global FA loss $\mathcal{L}_{\text{FA}}$ is formulated as
\begin{equation}
\begin{gathered}
    \mathcal{L}_{\text{FA}} = \lVert\boldsymbol{\mu}_{s} - \boldsymbol{\hat{\mu}}_{t}\rVert_2^2 + \lVert\boldsymbol{\Sigma}_{s} - \boldsymbol{\hat{\Sigma}}_{t}\rVert_F^2,
\end{gathered}
\end{equation}
where $\boldsymbol{\mu}_s, \boldsymbol{\Sigma}_s$ denote the mean and covariance of source features without considering class information, and $\boldsymbol{\hat{\mu}}_t, \boldsymbol{\hat{\Sigma}}_t$ indicate the mean and covariance estimated from a mini-batch of test samples. $\lVert\cdot\rVert_2, \lVert\cdot\rVert_F$ denote the Euclidean norm and Frobenius norm, respectively.
As shown in Fig.~\ref{fig:motivation} (a) (red lines), while the Global FA drastically decreases the intra-class distance, it accompanies a significant reduction of inter-class distance which needs to be high for achieving a reasonable level of image classification accuracy.
Fig.~\ref{fig:motivation} (b) (red line) also verifies such a point by visualizing the degraded image classification accuracy.

To address such a problem, we take the \emph{class information} into account when aligning features.
Aligning the source and target distributions in a class-wise manner would be one straightforward approach.
However, at test time, there exist very few samples for each class in a mini-batch to precisely estimate class-conditional distributions of test data.
Thus, we align individual test samples to the source class-conditional distribution of the predicted classes by using the Mahalanobis distance.
Note that we adopt the predicted class of each sample as a proxy of its ground truth label since we do not have access to the true label~\cite{pl}.
Specifically, with the given source class-conditional Gaussian distributions $\mathcal{N}(\boldsymbol{\mu}_c, \boldsymbol{\Sigma}_c)$, the loss $\mathcal{L}_{\text{intra}}$ minimizing the intra-class distance is defined as
\begin{equation}
\begin{gathered}
\mathcal{L}_{\text{intra}} = \frac{1}{N}\sum_{n=1}^{N}D_{\text{intra}}(x_{t, n}, \hat{y}_{t, n}),
\end{gathered}
\end{equation}
where $\hat{y}_{t, n}$ indicates the predicted class of the target sample $x_{t, n}$, and $N$ denotes the number of target samples.
While utilizing $\mathcal{L}_{\text{intra}}$ effectively reduces the intra-class distance, it still decreases the inter-class distance as shown in Fig.~\ref{fig:analysis} (a) (green line). We present our method in the next section that addresses this issue by adopting the loss function to also enlarge the inter-class distance. 

\subsection{CAFA: Class-Aware Feature Alignment}
\label{subsec:method_CAFA}
\noindent\textbf{Pre-calculation of source statistics}
As shown in the pre-stage of Fig.~\ref{fig:method_overview}, we calculate $C$ class-conditional Gaussian distributions $P(g_s(x_s)|y=c) = \mathcal{N}(g_s(x_s)|\boldsymbol{\mu}_c, \boldsymbol{\Sigma}_c)$ with the pretrained feature extractor $g_s(\cdot)$ over source training samples $(x_s, y_s)$ with the following equations:
\begin{equation}
\begin{gathered}
    \boldsymbol{\mu}_c = \frac{1}{N_c}\sum_{n=1}^{N_c}g_s(x_{s, n_c}),\\
    \boldsymbol{\Sigma}_c = \frac{1}{N_c}\sum_{n=1}^{N_c}(g_s(x_{s, n_c}) - \boldsymbol{\mu}_c)(g_s(x_{s, n_c}) - \boldsymbol{\mu}_c)^\top,
\end{gathered}
\end{equation}
where $N_c$ denotes the number of training samples of class $c$, and $x_{s, n_c}$ indicates training samples of class $c$.

\noindent\textbf{Test-time adaptation}
With the source class-conditional distributions $P(g_s(x_s)|y=c)$, we perform class-aware feature alignment at test time as illustrated in the test-time adaptation stage of Fig.~\ref{fig:method_overview}.
We initialize a model using the weights of pretrained networks $f_s(\cdot)$ and perform adaptation to target data considering both intra- and inter-class distances.
Our final loss $\mathcal{L}_{\text{CAFA}}$ is defined as
\begin{equation}
    \mathcal{L}_{\text{CAFA}} = \frac{1}{N}\sum_{n=1}^{N}\log\frac{D_{\text{intra}}(x_{t, n}, \hat{y}_{t, n})}{\sum_{c=1}^{C}D(x_{t, n}; \boldsymbol{\mu}_c, \boldsymbol{\Sigma}_c)}.
\end{equation}
As shown in Fig.~\ref{fig:analysis} (a) (blue line), our final loss aligns the source and target distributions in a desirable way by reducing the intra-class distance and enlarging the inter-class distance.

\begin{table*}[ht!]
\begin{center}
\footnotesize
\scalebox{0.85}{
\begin{tabular}{c|ccccccccccccccc|c}
\toprule
Method  & Gaus. & Shot & Impu. & Defo. & Glas. & Moti. & Zoom & Snow & Fros. & Fog & Brig. & Cont. & Elas. & Pixe. & Jpeg. & Average \\
\drule
Source & 48.73 & 44.00 & 57.00 & 11.84 & 50.78 & 23.38 & 10.84 & 21.93 & 28.24 & 29.41 & 7.01 & 13.27 & 23.38 & 47.88 & 19.46 & 29.14 \\
BN & 17.34 & 16.36 & 28.25 & 9.89 & 26.11 & 14.27 & 8.15 & 16.29 & 13.82 & 20.69 & 8.58 & 8.49 & 19.67 & 11.74 & 14.17 & 15.59 \\
PL & 17.22 & 16.07 & 27.85 & 9.74 & 25.94 & 14.13 & 8.07 & 16.12 & 13.78 & 20.14 & 8.53 & 8.53 & 19.73 & 11.65 & 13.94 & 15.43 \\
FR-Online$^\dagger$ & 17.23 & 16.15 & 27.31 & 10.07 & 25.58 & 14.12 & 8.35 & 16.17 & 13.67 & 20.01 & 8.64 & 8.65 & 19.48 & 11.82 & 14.20 & 15.43 \\
TFA-Online$^{\dagger}$ & 15.80 & 14.91 & 23.89 & 9.29 & 23.08 & 12.82 & 7.41 & 13.93 & 12.60 & 16.41 & 7.43 & 7.95 & 17.24 & 12.00 & 12.86 & 13.84 \\
TTT++-Online$^{\dagger}$ & 16.80 & 14.92 & 21.99 & 9.60 & 22.97 & 12.32 & 7.55 & 13.14 & 12.67 & 14.33 & 7.06 & 7.85 & 17.27 & 11.63 & 12.74 & 13.52\\
TENT$^\dagger$ & 15.95 & 14.55 & 24.72 & 9.03 & 23.25 & 12.74 & 7.47 & 13.91 & 12.78 & 16.66 & 8.13 & 8.12 & 18.30 & 10.85 & 13.21 & 13.98 \\
EATA$^\dagger$ & 16.73	& 15.42 & 25.09 & 9.83 & 24.10 & 13.36 & 8.45 & 15.02 & 13.64 & 17.39 & 8.63 & 8.44 & 19.08 & 11.70 & 13.97 & 14.72 \\
\midrule
\textbf{CAFA (Ours)} & \textbf{14.28} & \textbf{12.70} & \textbf{21.12} & \textbf{7.73} & \textbf{20.84} & \textbf{10.55} & \textbf{6.75} & \textbf{11.93} & \textbf{11.31} & \textbf{13.33} & \textbf{6.95} &	\textbf{7.13} & \textbf{16.08} & \textbf{9.59} & \textbf{11.67} & \textbf{12.13} \\
\toprule
Source & 80.77 & 77.84 & 87.75 & 39.62 & 82.26 & 54.22 & 38.38 & 54.58 & 60.19 & 68.11 & 28.86 & 50.93 & 59.54 & 72.27 & 49.96 & 60.35  \\
BN & 47.37 & 45.58 & 60.10 & 34.01 & 56.70 & 40.99 & 32.05 & 46.53 & 42.57 & 54.41 & 32.56 & 33.30 & 48.83 & 37.47 & 39.43 & 43.46 \\
PL & 46.74 & 45.26 & 59.21 & 33.83 & 56.08 & 40.29 & 31.64 & 46.10 & 42.07 & 53.74 & 32.24 & 33.08 & 48.24 & 37.11 & 39.01 & 43.00 \\
FR-Online$^\dagger$ & 47.16 & 45.60 & 59.85 & 34.09 & 56.70 & 41.06 & 32.20 & 46.44 & 42.65 & 54.37 & 32.72 & 33.48 & 48.85 & 37.49 & 39.45 & 43.47 \\ 
TFA-Online$^{\dagger}$ & 44.68 & 43.28 & 56.17 & 32.47 & 54.11 & 37.48 & 30.32 & 42.46 & 39.73 & 47.57 & 30.18 & 32.52 & 45.34 & 36.81 & 37.28 & 40.69 \\
TTT++-Online$^{\dagger}$ & 43.70 & 41.84 & 55.77 & 31.15 & 53.38 & 35.54 & 29.98 & 41.13 & 38.70 & 45.08 & 29.14 & 30.34 & 44.69 & 35.47 & 37.37 & 39.55 \\
TENT$^{\dagger}$ & 43.11 & 41.70 & 53.30 & 31.35 & 51.08 & 36.34 & 29.90 & 42.73 & 38.99 & 45.13 & 29.64 &	30.62 & 44.03 & 34.23 & 36.34 & 39.23 \\
EATA$^{\dagger}$ &  43.12 & 41.94 & 52.20 & 32.02 & 50.35 & 36.56 & 30.42 & 41.94 & 39.31 & 43.52 & 29.88 & 30.89 & 44.75 & 34.55 & 37.10 & 39.24 \\
\midrule
\textbf{CAFA (Ours)} & \textbf{41.60} & \textbf{39.77} & \textbf{50.45} & \textbf{30.17} & \textbf{48.35} & \textbf{34.65} & \textbf{28.76} & \textbf{39.52} & \textbf{37.42} & \textbf{41.25} & \textbf{27.95} & \textbf{29.54} & \textbf{42.37} & \textbf{32.87} & \textbf{35.02} & \textbf{37.31}\\
\bottomrule
\end{tabular}}
\vspace{0.1cm}
\caption{Classification error (\%) on the CIFAR10-C (upper group) and CIFAR100-C (lower group) datasets with severity level 5 corruptions.
$^\dagger$ denotes the results obtained from the official codes.}
\vspace{-0.6cm}
\label{tab:cifar}
\end{center}
\end{table*}

\begin{table}[ht!]
\begin{center}
\footnotesize
\begin{tabular}{l|c}
\toprule
Method  & Averaged Error (\%) $\downarrow$ \\
\drule
ResNet-26 (GroupNorm) & 32.70 \\
\,\, $\bullet$ MEMO~\cite{memo} & 29.68 \\
ResNet-26 (GroupNorm)$+$JT & 35.30 \\
\,\, $\bullet$ TTT~\cite{ttt} & 20.00 \\
\,\, $\bullet$ TTT (Episodic) & 32.85 \\
ResNet-26 (BatchNorm) & 34.93 \\
\,\, $\bullet$ FR-Online~\cite{bufr} & 18.43 \\
\,\, $\bullet$ TENT~\cite{tent} & 17.25 \\
\midrule
\,\, $\bullet$ \textbf{CAFA (Ours)} & \textbf{16.72} \\
\bottomrule
\end{tabular}
\vspace{0.2cm}
\caption{Classification error (\%) on the CIFAR10-C dataset with severity level 5 corruptions using ResNet-26 networks. All the numbers are obtained from the official codes. JT denotes joint training.}
\vspace{-0.6cm}
\label{tab:cifar10_R26}
\end{center}
\end{table}

\subsection{Theoretical Background}
\noindent\textbf{Gaussian assumption of features}
Recent studies~\cite{mahalanobis, covariance} present theoretical justifications about the Gaussian assumption of features when the network is trained with the Softmax function.
For image classification, a discriminative classifier is trained using the Softmax function whose posterior distribution is 
\begin{equation}
p(y=c|x) = \frac{\exp(w_c^\intercal x + b_c)}{\sum_{c^\prime}{\exp(w_{c^\prime}^\intercal x + b_{c^\prime})}},
\end{equation}
where $x, y$ denote input features and labels, and $w_c, b_c$ indicate weight and bias.
However, a generative classifier such as Gaussian discriminant analysis (GDA) can also be used for classification.
GDA defines posterior distribution by assuming that a class distribution follows the multivariate Gaussian distribution $p(x|y=c) = \mathcal{N}(x|\mu_c, \Sigma_c)$, and a class prior distribution follows the Bernoulli distribution $p(y=c) = \frac{\beta_c}{\sum_{c^\prime}{\beta_{c^\prime}}}$.
Additionally, GDA assumes all the class-conditional distributions share the same covariance, \textit{i.e.,} $\Sigma_c = \Sigma$.
The posterior distribution of GDA is represented as
\begin{align} 
\label{equ:gda_posterior_prob}
&p(y=c|x) = \frac{p(y=c)p(x|y=c)}{\sum_{c^\prime}{p(y=c^\prime)p(x|y=c^\prime)}}\\ \nonumber
&= \frac{\exp(\mu_c^\intercal\Sigma^{-1} x - \frac{1}{2}\mu_c^\intercal\Sigma^{-1}\mu_c + \log\beta_c)}{\sum_{c^\prime}\exp(\mu_{c^\prime}^\intercal\Sigma^{-1} x - \frac{1}{2}\mu_{c^\prime}^\intercal\Sigma^{-1}\mu_{c^\prime} + \log\beta_{c^\prime})}.
\end{align}
The posterior distribution of GDA becomes equivalent to the one from the Softmax function if we set the weight $w_c = \mu_c^\intercal\Sigma^{-1}$ and the bias $b_c = - \frac{1}{2}\mu_c^\intercal\Sigma^{-1}\mu_c + \log\beta_c$.
This derivation implies that source features $x$ may follow the Gaussian distribution when the network is trained with the Softmax function.

\noindent\textbf{Interpretation of CAFA}
Even though our loss is devised from intuition (Fig.~\ref{fig:motivation}), we can interpret our loss by using the posterior distribution of GDA.
By assuming a uniform prior distribution and assuming each class has its own covariance, the negative log-posterior can be simplified as
\begin{align}
    &-\log{p(y=c|x)} \\ \nonumber
    &= -\log{\frac{\exp({-\frac{1}{2}D(x; \mu_c, \Sigma_c) -\frac{1}{2}\log\lvert\Sigma_c\rvert)}}{\sum_{c^\prime}\exp({-\frac{1}{2}D(x; \mu_{c^\prime}, \Sigma_{c^\prime}) - \frac{1}{2}\log\lvert\Sigma_{c^\prime}\rvert)}}}.
\end{align}
While this term and our loss both play a similar role in minimizing/maximizing the intra-/inter-class distances, we observed that the above term generates large gradients since it can get easily saturated due to the high variance of Mahalanobis distances.
On the other hand, we empirically found that our final loss is more stable, and thus, allows us to use a higher learning rate, achieving state-of-the-art performance.
From these analyses, we believe our loss generalizes well if the source model is trained with the Softmax function.

\section{Experiments}
\label{sec:experiments}
This section first demonstrates evaluation results in two different settings: 1) robustness to corruptions, and 2) domain adaptation beyond image corruptions.
Then, we present in-depth analyses of our method by conducting ablation studies and providing visualizations of the representation space.

\noindent\textbf{Baselines}
For evaluations, we consider the following baselines in our experiments: no adaptation (Source), test-time normalization (BN)~\cite{norm}, pseudo label (PL)~\cite{pl}, test-time training (TTT)~\cite{ttt}, test-time entropy minimization (TENT)~\cite{tent}, efficient anti-forgetting test-time adaptation (EATA)~\cite{eata}, constrastive test-time-adaptation (AdaContrast)~\cite{adacontrast}, test-time template adjuster (T3A)~\cite{t3a}, marginal entropy minimization with one test-point (MEMO)~\cite{memo}, feature restoration (FR-Online)~\cite{bufr}, test-time feature alignment (TFA-Online)~\cite{ttt++}, and test-time training++ (TTT++-Online)~\cite{ttt++}\footnote{Online adaptation denotes predicting incoming test samples immediately, while offline adaptation predicts the test samples after several iterations of the entire test data.}.
For fair comparisons, we note that FR-Online, TFA-Online, TTT++-Online, and CAFA store the training statistics before deployment for test-time adaptation while others do not require such a step.
Further details about the baselines can be found in the supplementary materials.

\begin{table*}[t!]
\begin{center}
\footnotesize
\scalebox{0.9}{
\begin{tabular}{c|ccccccccccccccc|c}
\toprule
Method  & Gaus. & Shot & Impu. & Defo. & Glas. & Moti. & Zoom & Snow & Fros. & Fog & Brig. & Cont. & Elas. & Pixe. & Jpeg. & Average \\
\drule
Source & 94.86 & 93.14 & 97.25 & 86.96 & 88.45 & 77.87 & 77.63 & 83.71 & 80.35 & 92.85 & 81.94 & 98.55 & 69.03 & 58.89 & 55.64 & 82.47\\
BN & 68.74 & 67.80 & 74.72 & 68.71 & 77.13 & 61.04 & 59.71 & 67.37 & 67.21 & 76.52 & 66.09 & 93.73 & 61.32 & 55.07 & 55.75 & 68.06 \\
PL & 68.11 & 66.95 & 74.18 & 67.92 & 76.52 & 60.32 & 58.87 & 67.08 & 66.63 & 75.99 & 65.34 & 93.38 & 60.82 & 54.77 & 55.50 & 67.49 \\
TENT$^\dagger$ & 64.11 & 63.72 & 70.35 & 63.22 & 73.39 & 56.64 & 55.07 & 64.28 & 62.99 & 70.39 & 60.88 & \textbf{92.44} & 57.17 & 51.72 & 52.94 & 63.95 \\
EATA$^\dagger$ & 65.18 & 64.20 & 72.84 & 63.71 & 74.81 & 56.90 & 55.81 & 65.28 & 64.43 & 71.79 & 60.11 & 97.54 & 57.98 & 52.20 & 53.85 & 65.11\\
\midrule
\textbf{CAFA (Ours)} & \textbf{63.68} & \textbf{63.04} & \textbf{70.12} & \textbf{61.27} & \textbf{71.30} & \textbf{55.22} & \textbf{54.34} & \textbf{63.31} & \textbf{61.88} & \textbf{67.96} & \textbf{59.15} & 92.53 & \textbf{56.16} & \textbf{51.21} & \textbf{52.60} & \textbf{62.92} \\
\bottomrule
\end{tabular}}
\caption{Classification error (\%) on the TinyImageNet-C dataset with severity level 5.
$^\dagger$ denotes the results obtained from the official codes.}
\vspace{-0.6cm}
\label{tab:tinyimagenetc}
\end{center}
\end{table*}

\begin{table*}[t!]
\begin{center}
\footnotesize
\scalebox{0.9}{
\begin{tabular}{c|ccccccccccccccc|c}
\toprule
Method  & Gaus. & Shot & Impu. & Defo. & Glas. & Moti. & Zoom & Snow & Fros. & Fog & Brig. & Cont. & Elas. & Pixe. & Jpeg. & Average \\
\drule
Source & 97.79 & 97.07 & 98.15 & 82.08 & 90.18 & 85.22 & 77.50 & 83.11 & 76.69 & 75.57 & 41.07 & 94.57 & 83.05 & 79.39 & 68.35 & 81.99 \\
BN & 84.73 & 84.11 & 84.16 & 84.95 & 84.83 & 73.61 & 61.12 & 65.66 & 66.90 & 51.86 & 34.77 & 83.18 & 55.90 & 51.17 & 60.43 & 68.49 \\
PL & 85.66 & 88.98 & 84.34 & 93.01 & 91.91 & 65.85 & 54.46 & 55.59 & 69.07 & 43.48 & 32.75 & 98.85 & 47.67 & 41.88 & 48.61 & 66.81 \\
TENT$^\dagger$ & 71.26 & 69.54 & 69.99 & 71.95 & 72.85 & 58.77 & 50.69 & 52.86 & 58.89 & 42.57 & 32.68 & 73.29 & 45.17 & 41.57 & 47.94 & 57.33 \\
EATA$^\dagger$ & \textbf{65.00} &  \textbf{63.10} & \textbf{64.30} & \textbf{66.30} & \textbf{66.60} & \textbf{52.90} & \textbf{47.20} & \textbf{48.60} & \textbf{54.30} & \textbf{40.10} & \textbf{32.00} & \textbf{55.70} & \textbf{42.40} & \textbf{39.30} & \textbf{45.00} & \textbf{52.00}\\
\midrule
\textbf{CAFA (Ours)} & 69.59 & 67.29 & 68.03 & 71.09 & 70.87 & 56.13 & 50.03 & 50.77 & 56.77 & 41.86 & 33.24 & 61.30 & 43.76 & 40.87 & 47.03 & 55.24 \\
\bottomrule
\end{tabular}}
\caption{Classification error (\%) on the ImageNet-C dataset with severity level 5. 
$^\dagger$ denotes the results obtained from the official codes.}
\vspace{-0.7cm}
\label{tab:imagenet}
\end{center}
\end{table*}

\noindent\textbf{Implementation details}
We adopt the ResNet50~\cite{resnet} for our main experiments except for Table~\ref{tab:cifar10_R26}.
For the test-time adaptation, we set the batch size as 200 and utilize the Adam~\cite{adam} optimizer with a learning rate of 0.001 for adaptation. For ImageNet-C experiments, we adopt a batch size of 64, and we set a learning rate to 0.0025 for CAFA and 0.00025 for others with SGD optimizer.
Note that we only optimize the modulation parameters $\gamma, \beta$ in batch normalization layers, following Wang \textit{et al.}~\cite{tent}.

\subsection{Robustness to Corruptions}
\noindent\textbf{Datasets}
For corruption datasets, we evaluate methods on the CIFAR10-C, CIFAR100-C, TinyImageNet-C, and ImageNet-C~\cite{corruption} datasets.
CIFAR10~\cite{cifar} and CIFAR100~\cite{cifar} include 50,000 training samples and 10,000 test samples with 10 and 100 classes, respectively.
TinyImageNet~\cite{tinyimagenet} is a subset of the original ImageNet~\cite{imagenet} dataset, containing 100,000 training images and 10,000 validation images with 200 classes.
ImageNet~\cite{imagenet} contains 1.2 million training samples and 50,000 validation samples with 1,000 object categories.
CIFAR10-C and CIFAR100-C~\cite{corruption} datasets contain 15 different corruptions, and the corruptions are applied to the test set of CIFAR10 and CIFAR100 datasets.
Analogous to the CIFAR10-C / CIFAR100-C datasets, TinyImageNet-C and ImageNet-C datasets are composed of 15 corruption types, where the corruptions are applied to the validation set of TinyImageNet and ImageNet, respectively.
\begin{table*}[ht!]
\begin{center}
\footnotesize
\scalebox{0.9}{
\begin{tabular}{c|cccccccccccc|c}
\toprule
Method  & Ar$\rightarrow$Cl & Ar$\rightarrow$Pr & Ar$\rightarrow$Re & Cl$\rightarrow$Ar & Cl$\rightarrow$Pr & Cl$\rightarrow$Re & Pr$\rightarrow$Ar & Pr$\rightarrow$Cl & Pr$\rightarrow$Re & Re$\rightarrow$Ar & Re$\rightarrow$Cl & Re$\rightarrow$Pr & Average \\
\drule
Source & 67.33 & 46.68 & 36.81 & 68.52 & 54.22 & 53.91 & 66.25 & 71.25 & 40.76 & 45.16 & 65.02 & 29.65 & 53.80  \\
BN & 63.34 & 47.31 & 36.29 & 66.09 & 57.85 & 54.12 & 62.22 & 68.93 & 39.29 & 46.27 & 60.73 & 30.19 & 52.72 \\
PL & 63.18 & 46.25 & 35.87 & 65.68 & 56.14 & 53.16 & 61.80 & 68.34 & 38.51 & 45.45 & 60.53 & 29.62 & 52.04 \\
TENT$^{\dagger}$ & 61.47 & 44.33 & 34.82 & 62.75 & 52.22 & 49.16 & 61.60 & 66.19 & 36.26 & 44.66 & 58.08 & 28.14 & 49.97 \\
T3A & 62.29 & \textbf{41.41} & 34.11 & 64.65 & 51.05 & 48.91 & 60.61 & 66.96 & \textbf{35.69} & 45.69 & 59.22 & \textbf{27.64} & 49.85 \\
AdaContrast & 61.97 & 41.95 & 34.59 & 62.46 & 50.96 & 49.71 & \textbf{59.21} & 65.27 & 36.93 & 46.11 & 56.13 & 28.23 & 49.46 \\
EATA$^\dagger$ & 62.86 & 43.64 & 34.43 & 63.37 & \textbf{50.91} & 48.70 & 60.77 & 65.91 & 35.99 & 43.30 & 56.40 & 27.84 & 49.43 \\
\midrule
\textbf{CAFA (Ours)} & \textbf{59.73} & 42.64 & \textbf{34.01} & \textbf{61.39} & 51.23 & \textbf{47.69} & 60.28 & \textbf{63.92} & 35.87 & \textbf{42.89} & \textbf{54.91} & 27.84 & \textbf{48.53} \\
\bottomrule
\end{tabular}}
\vspace{+0.1cm}
\caption{Classification error (\%) on the OfficeHome~\cite{officehome} dataset.
$^\dagger$ denotes the results obtained from the official codes.}
\vspace{-0.8cm}
\label{tab:officehome}
\end{center}
\end{table*}

\begin{table}[ht!]
\begin{center}
\footnotesize
\scalebox{0.88}{
\begin{tabular}{c|cccccc|c}
\toprule
Method  & Clip. & Info. & Pain. & Quic. & Real & Sket. & Avg. \\
\drule
Source & 76.53 & 75.38 & 74.29 & 96.69 & 73.16 & 75.18 & 78.54  \\
BN & 76.14 & 79.25 & 72.90 & 93.15 & 74.17 & 68.80 & 77.40 \\
PL & 75.36 & 78.02 & 72.47 & 93.01 & 73.21 & 68.17 & 76.71 \\
TENT$^{\dagger}$ & 84.59 & 75.66 & 71.81 & 92.86 & 71.60 & 67.85 & 75.73 \\
T3A & 74.10 & 76.50 & 71.68 & 92.89 & 73.50 & 66.64 & 75.88 \\
EATA$^{\dagger}$ & 73.88 & 75.26 & 71.16 & 92.37 & 70.69 & 66.86 & 75.04\\
\midrule
\textbf{CAFA (Ours)} & \textbf{73.17} & \textbf{74.69} & \textbf{71.05} & \textbf{92.49} & \textbf{69.96} & \textbf{66.53} & \textbf{74.65} \\
\bottomrule
\end{tabular}}
\vspace{0.1cm}
\caption{Classification error (\%) on the DomainNet~\cite{domainnet} dataset.
$^\dagger$ denotes the results obtained from the official codes.}
\vspace{-0.6cm}
\label{tab:DomainNet}
\end{center}
\end{table}

\noindent\textbf{Quantitative evaluation and comparisons}
To evaluate the robustness to corruption, we utilize the pretrained networks on CIFAR10, CIFAR100, TinyImageNet, and ImageNet datasets and adapt the pretrained networks to their corresponding corruption datasets, respectively.

Table~\ref{tab:cifar} shows the image classification errors (\%) on diverse corruption types with the severest corruption level in the CIFAR10-C and CIFAR100-C datasets.
As shown, our proposed method outperforms the baselines on all types of corruptions in both CIFAR10-C and CIFAR100-C datasets by a large margin.
Additionally, we conduct an experiment on the CIFAR10-C dataset with the ResNet-26 architecture to make further comparisons of CAFA with other test-time adaptation methods.
Note that MEMO~\cite{memo} and TTT~\cite{ttt} adopt ResNet-26 with group normalization layers as their base architecture.
As reported in Table~\ref{tab:cifar10_R26}, CAFA achieves the lowest error among the methods sharing the same pretrained model (\textit{i.e.,} ResNet-26 (BatchNorm)).
Furthermore, CAFA achieves the largest performance gain over the source model compared to MEMO~\cite{memo} and TTT~\cite{ttt}.

We further evaluate our method on a larger dataset, TinyImageNet-C, as shown in Table~\ref{tab:tinyimagenetc}.
Similar to the results of CIFAR10-C and CIFAR100-C, CAFA outperforms the baselines in all types of corruptions except for the contrast corruption on the TinyImageNet-C dataset.
Even in such a case, the increased error is 0.1\% which is marginal considering the performance gains in other corruption types.
Finally, we validate our method on the most challenging corruption dataset, ImageNet-C.
As reported in Table~\ref{tab:imagenet}, even though our method is not a top performer, it outperforms most of the baselines by a large margin, which is 26.8\% over the source model and 2.1\% over TENT.

In the CIFAR10-C and CIFAR100-C datasets, CAFA achieves lower error rates compared to the TFA-Online and FR-Online methods which align source and target distributions without considering the class information.
Such results demonstrate that considering both intra- and inter-class distances is important when performing feature alignments for test-time adaptation.
Note that TFA-Online and TTT++-Online methods are online adaptation methods based on the original TFA and TTT++~\cite{ttt++} algorithms.
Those are designed for offline adaptation that iterates multiple epochs over the entire set of test samples and predicts the test samples at once after multiple epochs.
In their offline adaptation setting, TFA and TTT++ achieve 11.87\% and 9.60\% error rates on the CIFAR10-C dataset.\footnote{Those numbers are obtained from Liu \textit{et al.}~\cite{ttt++}\vspace{-0.8cm}}

\subsection{Domain Adaptation beyond Image Corruptions}
This section presents the experimental results for domain adaptation datasets beyond image corruption.

\noindent\textbf{Datasets}
We adopt Office-Home~\cite{officehome} and DomainNet~\cite{domainnet} datasets, which are widely used domain adaptation datasets.
Office-Home~\cite{officehome} dataset consists of around 15,500 images and contains 65 categories of everyday objects with four distinct domains: Artistic images (Ar), Clip art images (Cl), Product images (Pr), and Real-world images (Re).
DomainNet~\cite{domainnet} dataset is the largest domain adaptation dataset containing around 0.6 million images of 345 categories on six different domains which are clipart, infograph, painting, quickdraw, real, and sketch.

\noindent\textbf{Quantitative evaluation and comparisons}
We evaluate our method on 12 different adaptation scenarios of the Office-Home~\cite{officehome} dataset, pretraining a model on one source domain and adapting it to the other domains.
Table~\ref{tab:officehome} shows the image classification errors (\%) on different adaptation scenarios of the OfficeHome~\cite{officehome} dataset.
As shown, CAFA consistently outperforms the baselines by a large margin.
Ours reduces the classification error of the source model by around 5.3\%, and that of TENT and EATA by around 1.4\% and 0.9\%, respectively.

Moreover, we also make comparisons with baselines on the DomainNet dataset, which is the largest domain adaptation dataset. For the experiment, we follow a similar evaluation protocol to the Office-Home dataset, \textit{i.e.,} pretraining a model on each source domain and adapting the pretrained model to the other five domains. Each column denotes the source domain, and the numbers are the averaged classification error of the other five domains. As shown in Table~\ref{tab:DomainNet}, CAFA outperforms the existing methods by 1.1\% and 0.4\% over TENT and EATA, respectively.

\begin{figure}[h!]
\centering
  \includegraphics[width=1.02\linewidth]{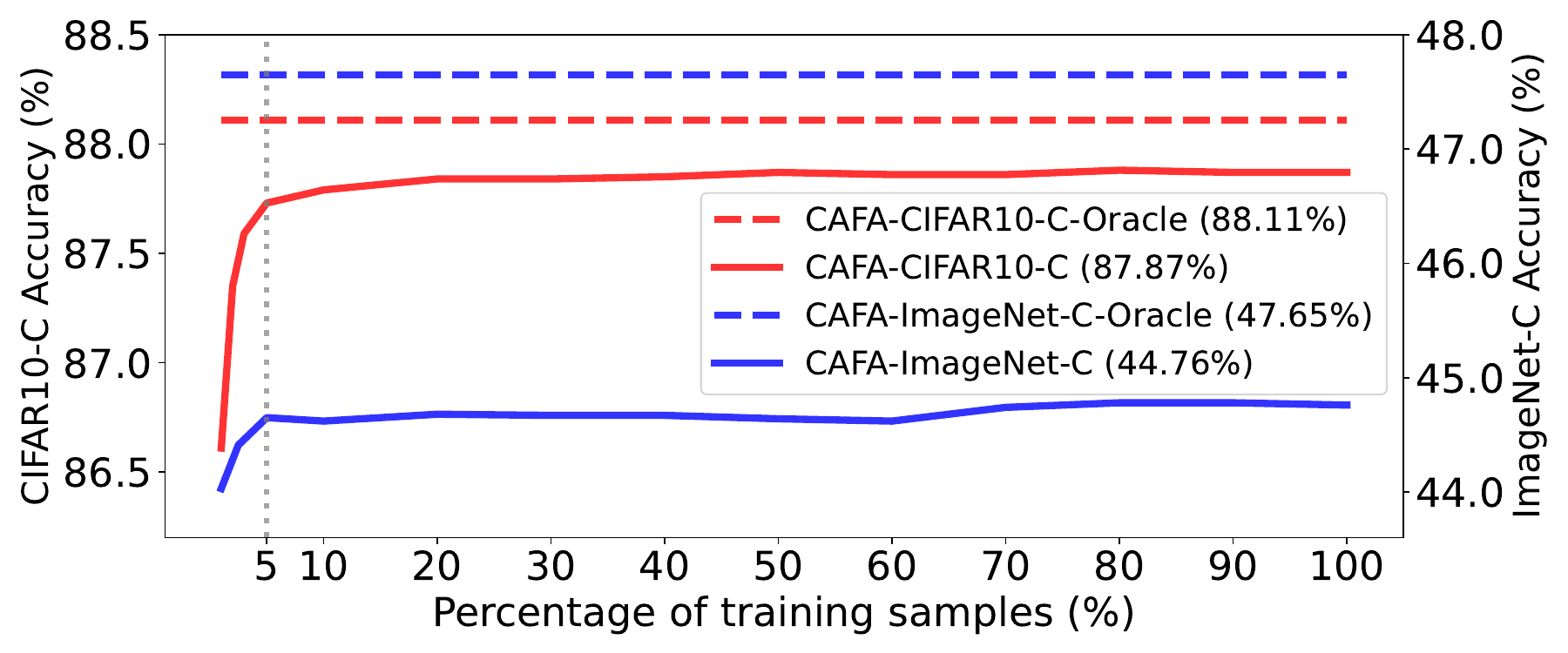}
  \vspace{-0.4cm}
    \caption{The test-time adaptation performance of CAFA on the CIFAR10-C and ImageNet-C datasets along with the percentage of train samples for calculating the train statistics.}
  \vspace{-0.4cm}
    \label{fig:train_sample}
\end{figure}

\subsection{Analysis}
\label{subsec:analysis}

\noindent\textbf{Pre-calculation of source statistics}
As aforementioned, CAFA requires access to the training samples to compute source statistics before adaptation.
We demonstrate that we can obtain robust statistics for adaptation even with a small number of training samples, outperforming the existing methods.
In Fig.~\ref{fig:train_sample}, the red lines denote the performance of CAFA (solid line) and CAFA-Oracle (dotted line) on the CIFAR10-C dataset, and the blue lines indicate the performance on the ImageNet-C dataset.
Note that oracle performances (\textit{i.e.,} applying CAFA with ground truth labels) are measured using all training samples for obtaining statistics.
As shown, around 5\% of the training samples are sufficient to pre-calculate robust source statistics for achieving high adaptation performance in both CIFAR10-C and ImageNet-C datasets.

\begin{table}[ht!]
\begin{center}
\footnotesize
\scalebox{0.9}{
\begin{tabular}{cc|cc|cc}
\toprule
\multicolumn{2}{c|}{\shortstack{Effectiveness of\\Intra-/Inter-Class Dist.}} & \multicolumn{2}{c|}{\shortstack{Updating Batch Norm.\\vs Full Parameters}} & \multicolumn{2}{c}{\shortstack{Tied vs Class-wise\\Covariance}} \\
\drule
Source & 29.14 & Source & 29.14 & Source & 29.14 \\
Global FA & 19.12 & CAFA-Full & 12.66 & CAFA-Tied & 12.47 \\
Intra-Class Dist. & 13.02 & \textbf{CAFA} & \textbf{12.13} & \textbf{CAFA} & \textbf{12.13} \\
\textbf{CAFA} & \textbf{12.13} & & & & \\
\bottomrule
\end{tabular}
}
\vspace{+0.1cm}
\caption{Our ablation results on the CIFAR10-C dataset.}
\vspace{-0.5cm}
\label{tab:ablation}
\end{center}
\end{table}

\begin{table}[ht!]
\vspace{-0.4cm}
\begin{center}
\footnotesize
\scalebox{0.78}{
\begin{tabular}{c|cc|cc|cc}
\toprule
&\multicolumn{2}{c|}{Intra-/Inter-Class} & \multicolumn{2}{c|}{BN vs Full Param.} & \multicolumn{2}{c}{Tied vs Class-wise Cov.} \\
\drule
\multirow{3}{*}{CIFAR10-C}& Global FA & 19.12 & Source & 29.14 & Source & 29.14 \\
& Intra-Class & 13.02 & CAFA-Full & 12.66 & CAFA-Tied & 12.47 \\
& \textbf{CAFA} & \textbf{12.13} & \textbf{CAFA} & \textbf{12.13} & \textbf{CAFA} & \textbf{12.13} \\
\midrule
\multirow{3}{*}{ImageNet-C} & Global FA & 72.38 & Source & 81.99 & Source & 81.99 \\
& Intra-Class & 59.12 & \textbf{CAFA-Full} & \textbf{55.24} & CAFA-Tied & 56.00 \\
& \textbf{CAFA} & \textbf{55.24} & \textbf{CAFA} & \textbf{55.24} & \textbf{CAFA} & \textbf{55.24} \\
\midrule
\multirow{3}{*}{Office-Home} & Global FA & 55.35 & Source & 53.80 & Source & 53.80 \\
&Intra-Class & 49.50 & CAFA-Full & 48.47 & CAFA-Tied & 48.53 \\
&\textbf{CAFA} & \textbf{48.27} & \textbf{CAFA} & \textbf{48.27} & \textbf{CAFA} & \textbf{48.27} \\
\bottomrule
\end{tabular}
}
\vspace{+0.1cm}
\caption{Our ablation results on the CIFAR10-C, ImagNet-C, and Office-Home datasets.}
\label{tab:ablation}
\vspace{-0.4cm}
\end{center}
\end{table}

\noindent\textbf{Effectiveness of Intra-/Inter-Class Distance}
To further validate our motivation for considering intra- and inter-class distances, we conduct ablation studies on the CIFAR10-C, ImageNet-C, and Office-Home datasets.
As reported in the left group of Table~\ref{tab:ablation}, Global FA performs poorly in TTA since it does not consider the class information.
In the case of reducing the intra-class distance only, it improves the classification errors over the source model, which is also effective.
However, considering both intra- and inter-class distances (CAFA) achieves the lowest classification errors.  
Such results demonstrate our initial intuition is valid, which is elaborated on in Section~\ref{sec:introduction}.
Note that we obtain consistent results on the CIFAR100-C dataset, as presented in the supplementary material.

\begin{figure*}[t!]
\centering
  \includegraphics[width=0.9\linewidth]{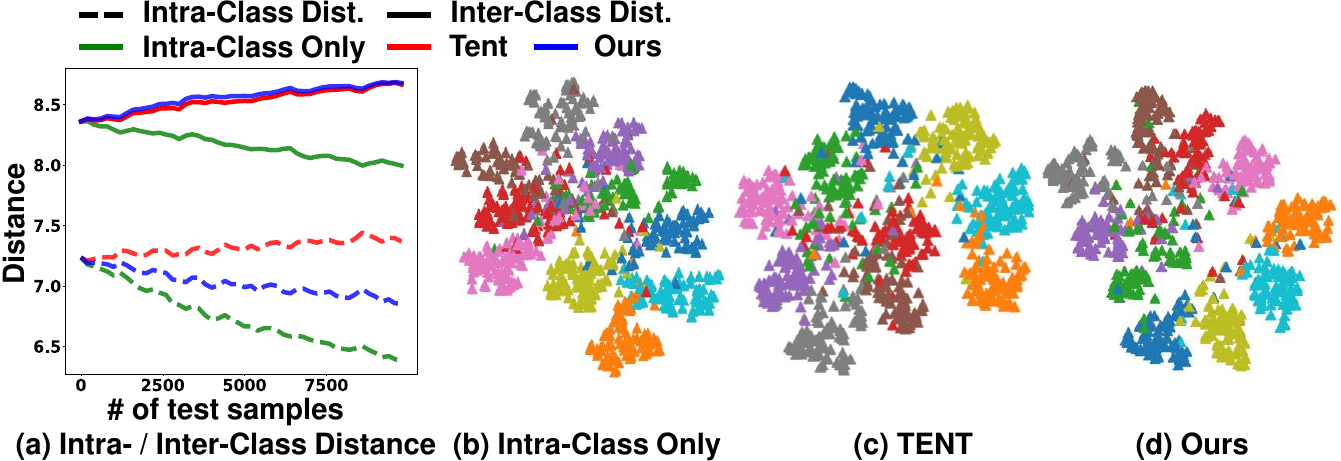}
  \vspace{0.1cm}
    \caption{(a) illustrates the change of intra-class distance (dotted lines) and inter-class distance (solid line) as adaptation proceeds.
    (b-d) show the t-SNE visualizations of applying (b) intra-class distance only, (c) TENT~\cite{tent}, and (d) ours.
    All visualizations are obtained from the Gaussian noise corruption in the CIFAR10-C dataset.}
  \vspace{-0.6cm}
    \label{fig:analysis}
\end{figure*}

\noindent\textbf{Updating the entire parameters of feature extractor}
In our main experiments, we only update the modulation parameters $\beta, \gamma$ of batch normalization layers in the networks, following Wang \textit{et al.}~\cite{tent}.
In this ablation study, we further validate CAFA by updating the entire parameters of the feature extractor.
Note that the classifier $h(\cdot)$ cannot be updated by our loss since CAFA performs alignments at a feature level.
CAFA-Full in the middle group of Table~\ref{tab:ablation} shows the result of updating the full parameters when utilizing CAFA.
While it shows superior performance compared to the baselines, updating the batch normalization layers outperforms the case of updating the full parameters except for the ImageNet-C dataset.
In the case of the ImageNet-C dataset, no difference in their performance is observed.
As pointed out in Wang \textit{et al.}~\cite{tent}, updating the full model may cause the model to diverge from what they learned from training.
Furthermore, we conjecture that the number of samples during test-time adaptation may be not sufficient to optimize the entire parameters to converge.

\noindent\textbf{Impact of using tied covariance}
We adopt the class-wise covariance matrices for CAFA in the main experiments. 
However, in Gaussian discriminant analysis (GDA), it is assumed that all the class-conditional Gaussian distributions share the same covariance matrix (\textit{i.e.,} tied covariance).
We conduct an ablation study regarding such an issue in the right group of Table~\ref{tab:ablation}.
We observe that CAFA with tied covariance shows less performance improvement than utilizing the class-wise covariances.
We conjecture that it is because class-wise covariances represent the statistics of each class of the source distributions more precisely than the tied covariance.
Moreover, as pointed out in Lee \textit{et al.}~\cite{covariance}, deep networks are not trained to share the same covariance matrix for all class-conditional distributions.
However, regardless of the covariance types, our method still outperforms the existing baselines.

\noindent\textbf{Effectiveness of Mahalanobis distance}
To provide concrete background on the choice of Mahalanobis distance, we conduct experiments along with different distance or divergence types as in Table~\ref{tab:distance_type}.
As shown in the CIFAR10-C experiments, CAFA achieves the best accuracy while class-aware KL-divergence also improves the source performance by a large margin.
However, it is not applicable to the ImageNet-C dataset since each test batch does not have enough samples to estimate the distribution for each class.
On the other hand, Mahalanobis distance robustly outperforms other types since it measures the distance between a distribution and a sample.
\begin{table}[ht!]
\begin{center}
\footnotesize
\begin{tabular}{c|c|c|c}
\toprule
\cmidrule{2-4}
Alignment types & Methods & CIFAR10-C & ImageNet-C\\
\drule
\multirow{2}{*}{Class agnostic} & Global FA & 19.12 & 72.38 \\
& KL-divergence & 13.23 & N/A \\
\midrule
\multirow{3}{*}{Class aware} & KL-divergence & 12.53 & N/A \\
& Euclidean & 13.20 & 56.30 \\
& CAFA & \textbf{12.13} & \textbf{55.24} \\
\bottomrule
\end{tabular}
\vspace{+0.2cm}
\caption{Ablation study on different distance types. Note that Euclidean distance is a special case of Mahalanobis distance.}
\vspace{-0.7cm}
\label{tab:distance_type}
\end{center}
\end{table}

\begin{table}[ht!]
\begin{center}
\footnotesize
\scalebox{0.9}{
\begin{tabular}{c|c|ccc}
\toprule
\multicolumn{2}{c|}{Conditions} & TENT$^\dagger$ & CAFA (Ours)\\
\drule
\multirow{4}{*}{Batch Size}& BS $=8$ & 14.47 & 13.02 \\
& BS $=4$ & 14.52 & 13.05 \\
& BS $=2$ & 15.10 & 13.43 \\
& BS $=1$ & 17.66 & 15.00 \\
\midrule
\multicolumn{2}{c|}{Source} & \multicolumn{2}{c}{29.14} \\
\midrule
\multirow{6}{*}{Label Shifts} & $s=0.1$ & 14.92 & 13.32 \\
& $s=0.3$ & 15.80 & 13.40 \\
& $s=0.5$ & 17.15 & 13.83 \\
& $s=0.7$ & 19.68 & 14.73 \\
& $s=0.9$ & 23.70 & 16.47 \\
& $s=1.0$ & 26.01 & 17.61 \\
\midrule
\multicolumn{2}{c|}{Source} & \multicolumn{2}{c}{29.14} \\
\bottomrule
\end{tabular}}
\vspace{0.1cm}
\caption{Classification error (\%) on the CIFAR10-C dataset with severity level 5 corruptions with small batch sizes (upper group) and label shifts (lower group).}
\vspace{-0.6cm}
\label{tab:batch_label_ablation}
\end{center}
\end{table}

\noindent\textbf{Impact of small batch sizes and label shifts}
To show the wide applicability of CAFA to various deployment scenarios, we conduct ablation studies on 1) small batch sizes and 2) label shifts.
In the experiment on small batch sizes ($<$ 10), we show that test-time adaptation approaches that assume a batch of test instances can still be effective in such scenarios, even when the batch size is equal to 1.
To tackle this challenge, we modify how we compute the batch statistics.
As pointed out in Schneider \textit{et al}.~\cite{improving}, an effective solution is inferring test batch normalization statistics by leveraging the training batch statistics.
Based on such intuition, we apply Schneider \textit{et al}.~\cite{improving} to CAFA and TENT and measure the performance with batch sizes less than 10.
As reported in the upper group of Table~\ref{tab:batch_label_ablation}, CAFA and TENT maintain reasonable performance even when the batch size equals 1, improving the source model by 14.1\% and 11.5\% respectively.

Additionally, we conduct an experiment assuming the label distribution of each test batch changes (\textit{i.e.,} label shifts).
To simulate the label shift, we sample the test instances from the multinomial distribution and change the distribution for each batch.
With the given severity $s$ and the number of classes $c$, we change the multinomial distribution from $[s + \frac{1-s}{c}, \frac{1-s}{c}, ..., \frac{1-s}{c}]$ to $[\frac{1-s}{c}, \frac{1-s}{c}, ..., s + \frac{1-s}{c}]$, where each element of the array is the probability of the corresponding class being sampled. For example, when the severity $s=1$, the multinomial distribution changes from $[1, 0, 0, ..., 0]$ to $[0, 0, 0, ..., 1]$.
Based on the same intuition from the experiment regarding the batch sizes, we apply Schneider \textit{et al.}~\cite{improving} to CAFA and TENT methods.
As shown in the lower group of Table~\ref{tab:batch_label_ablation}, CAFA maintains reasonable performance even when the label shift happens at the most by improving the source model by 11.5\%.
TENT also improves the source model under all label shift severities, but the performance gains are less than CAFA.
We conjecture that this is because CAFA takes benefits of the feature alignment.
Adapting a model without alignments may result in divergence from the source distribution and may collapse to trivial solutions under severe label shifts.
On the other hand, in the case of CAFA, it aligns the target features to the source class-conditional distributions, which is more robust to such divergence.

\noindent\textbf{Visualizations}
To further validate our intuition, we visualize the change of the intra- and inter-class distances in Fig.~\ref{fig:analysis} (a): intra-class distance only (green line), TENT~\cite{tent} (red line), and CAFA (blue line).
As shown in Fig.~\ref{fig:analysis} (a), reducing the intra-class distance alone also decreases the inter-class distance, which harms the class-discriminability.
In the case of TENT~\cite{tent}, we observe that the intra-class distance does not decrease.
On the other hand, CAFA desirably reduces the intra-class distance while enlarging the inter-class distance.
Such improved class-discriminability is also shown in the t-SNE visualization in Fig.~\ref{fig:analysis} (d).
Ours shows more well-separated representation space in a class-wise manner compared to other methods (Fig.~\ref{fig:analysis} (b) and (c)).

\vspace{-0.2cm}
\section{Discussion}
\vspace{-0.2cm}
In this work, we proposed a \emph{simple yet effective} feature alignment that considers both intra- and inter-class distances, noting the importance of considering them for test-time adaptation.
Most of the existing feature alignments are generally conducted along with training on the source data, which allows a model to learn target distributions in a class-discriminative manner.
However, in the case of test-time adaptation where access to the source data is prohibited during adaptation, a model does not have a chance to learn test features in such a manner.
Our simple feature alignment that considers both intra- and inter-class distances effectively addresses such a challenge as shown in our analyses and extensive experiments.
However, there still remains room for improvements: 1) better pseudo-labeling method and 2) better learnable parameters than BN layers.
We hope our work inspires the following researchers to investigate more effective test-time adaptation methods.

\noindent\textbf{Pseudo labeling}
One limitation of our work is that we resort to using the predicted labels when assigning a class to each test sample.
This could be problematic in the early phase of adaptation since the model encounters samples from different distributions, achieving low classification accuracy.
Along with our novel perspective to consider both intra- and inter-class distance, we believe that replacing such a pseudo-labeling approach with advanced methods would further boost the adaptation performance of CAFA.

\noindent\textbf{Finding better learnable parameters than BN layers}
In our work, we only updated the modulation parameters in batch normalization layers.
Despite its demonstrated effectiveness, batch normalization parameters only take less than 1\% of the model parameters.
If we can find better learnable model parameters, it may further improve the adaptation performance of our proposed loss CAFA.

\vspace{-0.2cm}
\section{Acknowledgement}
\vspace{-0.2cm}
This work was supported by Institute of Information \& communications Technology Planning \& Evaluation (IITP) grant funded by the Korea government (MSIT) (No.2021-0-02068, Artificial Intelligence Innovation Hub) and the National Research Foundation of Korea (NRF) grant funded by the Korea government (MSIT) (No. 2022R1A5A708390811 and 2022R1A2B5B02001913).

\section{Appendix}

This supplementary material presents details about experimental settings, additional experimental results, and visualizations of our method, which are not included in the main paper due to the page limit.
Section~\ref{supsec:implementation_details} elaborates on the details about baselines, and Section~\ref{supsec:experiments} describes additional experimental results including ablation studies.
Finally, Section~\ref{supsec:visualization} presents the t-SNE visualizations of CAFA and TENT~\cite{tent}.

\subsection{Baselines}
\label{supsec:implementation_details}

\noindent{\textbf{Baselines}} For fair comparisons, we consider the following baselines in our experiments:
\begin{itemize}
\vspace{-0.2cm}
\item Source: Evaluating the pretrained network on the test data without any adaptation.
\vspace{-0.2cm}
\item Test-time normalization (BN)~\cite{norm_test_batch} updates the batch normalization statistics~\cite{batch_norm} on test data at test time.
\vspace{-0.2cm}
\item Pseudo label (PL)~\cite{pl} utilizes the predicted label as a label for optimizing the main task loss during testing. To be specific, we optimize the model by minimizing the cross-entropy loss with the pseudo labels.
\vspace{-0.2cm}
\item Test-time entropy minimization (TENT)~\cite{tent} updates the modulation parameters (\textit{i.e.,} $\beta$ and $\gamma$) of batch normalization~\cite{batch_norm} layers in the network by minimizing entropy on test data.
\vspace{-0.2cm}
\item Efficient anti-forgetting test-time adaptation (EATA)~\cite{eata} applies efficient sample selection to filter out redundant samples for adaptation and regularizes important weights using the Fisher matrix to prevent catastrophic forgetting.
\vspace{-0.2cm}
\item Marginal entropy minimization with one test point (MEMO)~\cite{memo} adapts a model to a single test point using test-time augmentation and minimizing marginal entropy.
\vspace{-0.2cm}
\item Feature Restoration (FR-Online)~\cite{bufr} pre-calculates the approximated training feature distributions and adapts a model to test domains by aligning the target feature distributions to the pre-obtained training feature distributions.
\vspace{-0.2cm}
\item Test-time template adjuster (T3A)~\cite{t3a} calculates a pseudo-prototype vector for each class on test time and classifies images using the distance between each instance and pseudo-prototype vectors.
\vspace{-0.2cm}
\item Contrastive test-time adaptation (AdaContrast)~\cite{adacontrast} applies contrastive learning along with online refinement of pseudo labels during test time.
\vspace{-0.2cm}
\item Test-time training (TTT)~\cite{ttt} first pre-trains the model with a self-supervision loss (\textit{i.e.,} rotation prediction) and adapts the pretrained model to test domains by minimizing the self-supervision loss as a proxy of the main task loss function.
\vspace{-0.2cm}
\item Test-time feature alignment (TFA-Online)~\cite{ttt++} aligns the source and target distributions by matching the first- and second-order statistics of outputs from both the penultimate layer and self-supervised task branch.
\vspace{-0.2cm}
\item Test-time training++ (TTT++-Online)~\cite{ttt++} updates the model by jointly aligning the first- and second-order statistics between the source and target distributions and optimizing the proxy loss (\textit{i.e.,} contrastive loss for self-supervision task) at test time.
\end{itemize}


\subsection{Additional Experimental Results}
\label{supsec:experiments}

\begin{table}[ht!]
\begin{center}
\footnotesize
\begin{tabular}{c|cc}
\toprule
Methods & Inference Time (ms) & FPS \\
\drule
Source & 17.82 & 77.97 \\
BN & 17.45 & 57.31 \\
TENT & 40.47 & 24.71 \\
CAFA & 41.65 & 24.01 \\
\bottomrule
\end{tabular}
\vspace{0.2cm}
\caption{Measured inference time (ms) and FPS for each method. We measure the inference time using a NVIDIA A100 GPU with an image resolution of 224$\times$224. The number is an averaged value of 300 trials.}
\vspace{-0.7cm}
\label{suptab:inference}
\end{center}
\end{table}

\subsubsection{Inference Time}
To show the efficiency of CAFA, we measure the inference time on a single NVIDIA A100 GPU and average the inference time over 300 trials with an image resolution of $224\times224$ in Table~\ref{suptab:inference}.
Our model shows around 24.01 frames per second (FPS).
Since CAFA and TENT optimize model parameters by minimizing the loss using gradient descent, those methods have a longer inference time than Source and BN methods.
We believe that such a result demonstrates that CAFA not only improves the test time adaptation performance but also maintains a reasonable level of efficiency.

\subsubsection{Further Ablation Studies}

\begin{table}[ht!]
\begin{center}
\footnotesize
\scalebox{0.92}{
\begin{tabular}{cc|cc|cc}
\toprule
\multicolumn{2}{c|}{\shortstack{Effectiveness of\\Intra-/Inter-Class Dist.}} & \multicolumn{2}{c|}{\shortstack{Updating Batch Norm.\\vs Full Parameters}} & \multicolumn{2}{c}{\shortstack{Tied vs Class-wise\\Covariance}} \\
\drule
Source & 29.14 & Source & 29.14 & Source & 29.14 \\
Global FA & 19.12 & CAFA-Full & 12.66 & CAFA-Tied & 12.47 \\
Intra-Class Dist. & 13.02 & \textbf{CAFA} & \textbf{12.13} & \textbf{CAFA} & \textbf{12.13} \\
\textbf{CAFA} & \textbf{12.13} & & & & \\
\midrule
Source & 60.35 & Source & 60.35 & Source & 60.35 \\
Global FA & 51.41 & CAFA-Full & 38.31 & CAFA-Tied & 38.19 \\
 Intra-Class Dist. & 41.51 & \textbf{CAFA} & \textbf{37.31} & \textbf{CAFA} & \textbf{37.31} \\
\textbf{CAFA} & \textbf{37.31} & & & & \\
\bottomrule
\end{tabular}
}
\vspace{0.2cm}
\caption{Our ablation results on the CIFAR10-C (upper group) and CIFAR100-C (lower group) datasets. The left group shows the effectiveness of considering intra- and inter-class distances, the middle group presents the comparison of updating the batch normalization parameters and full parameters in the feature extractor, and the right group describes the effects of using a tied covariance.}
\vspace{-0.5cm}
\label{suptab:ablation_extended}
\end{center}
\end{table}

\noindent{\textbf{Ablation studies on the CIFAR100-C dataset}} Along with the ablation studies in the main paper, we apply the same variants of our methods to the CIFAR100-C dataset.
Table~\ref{suptab:ablation_extended} shows the ablation studies on both CIFAR10-C (upper group) and CIFAR100-C (lower group) datasets.
Overall, we observe the similar results to the ablation studies in our main paper.
\begin{table}[h!]
\begin{center}
\footnotesize
\begin{tabular}{c|c|c|c}
\toprule
\multirow{2}{*}{Methods} & \multicolumn{3}{c}{Classification Error (\%)}\\
\cmidrule{2-4}
& CIFAR10-C & CIFAR100-C & ImageNet-C \\
\drule
Source & 29.14 & 60.35 & 81.99 \\
CAFA-Variance & 12.46 & 37.46 & 55.47 \\
\textbf{CAFA} & \textbf{12.13} & \textbf{37.31} & \textbf{55.24} \\
\bottomrule
\end{tabular}
\vspace{0.2cm}
\caption{Effectiveness of using variance in the CIFAR10-C, CIFAR100-C, and ImageNet-C datasets.}
\vspace{-0.5cm}
\label{suptab:ablation_variance_oracle}
\end{center}
\end{table}

\noindent\textbf{Effectiveness of using variance} Another variant of our loss is using variances of each class-conditional Gaussian distribution instead of using the full covariance.
While using variances does not consider the covariance between feature dimensions, it is still distinct for each class-conditional Gaussian distribution.
As reported in the upper group of Table~\ref{suptab:ablation_variance_oracle}, adopting class-wise variances also improve the source model significantly.
That is, variances can represent the class-conditional Gaussian distributions reasonably well.
However, using full covariance reaches the top performance in all three datasets.


\subsubsection{Different Corruption Severity Levels}
We compare CAFA with the baselines on the CIFAR10-C and CIFAR100-C datasets with different severity levels.
As reported in Tables~\ref{suptab:cifar_lv4}-\ref{suptab:cifar_lv1}, CAFA outperforms baselines on all severity levels of the CIFAR10-C and CIFAR100-C datasets.
Furthermore, we present the classification errors along with independent trials using different random seeds in Table~\ref{suptab:cifar_randomseeds}.
As shown, CAFA shows minimal performance variation considering the small standard deviation.

\subsection{Visualizations}
\label{supsec:visualization}
\begin{figure*}[ht!]
\vspace{-0.3cm}
  \includegraphics[width=\linewidth]{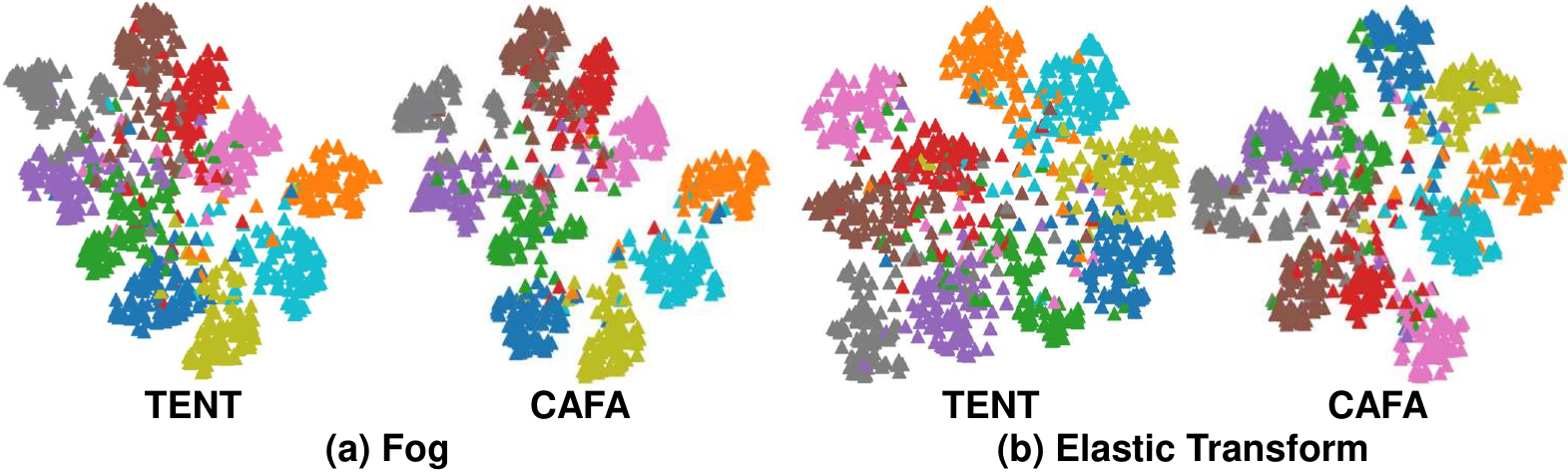}
\vspace{-0.1cm}
    \caption{t-SNE visualizations of ours (CAFA) and TENT from (a) Fog and (b) Elastic Transform corruptions with severity level 5 in the CIFAR10-C dataset.}
    \label{supfig:vis_corruptions}
\vspace{-0.3cm}
\end{figure*}

\begin{figure*}[hb!]
\vspace{-0.3cm}
  \includegraphics[width=\linewidth]{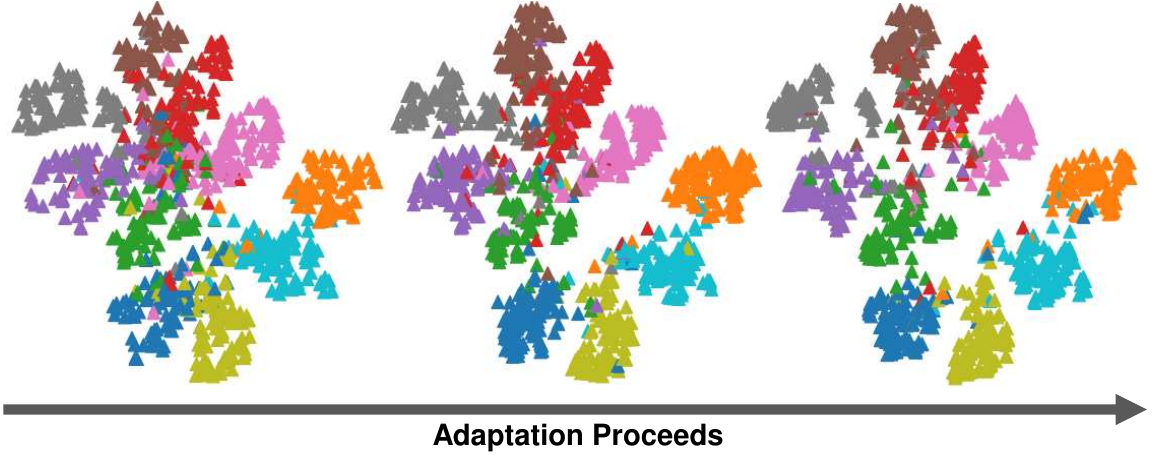}
\vspace{-0.3cm}
    \caption{Change of the representation space of test samples from our method as adaptation proceeds.
    Representation space is visualized by the t-SNE algorithm, and visualizations are obtained from the Fog corruption with severity level 5 in the CIFAR10-C dataset.
    }
    \label{supfig:vis_iteration}
\vspace{-0.3cm}
\end{figure*}

For the qualitative analysis, we visualize the representation space of test samples from our method and TENT~\cite{tent} by using the t-SNE algorithm.
Fig.~\ref{supfig:vis_corruptions} shows the t-SNE results on different corruption types in the CIFAR10-C dataset, and Fig.~\ref{supfig:vis_iteration} illustrates the change of representation space of test samples from our method as adaptation proceeds.
As shown, the representations of test samples are well-separated, and they desirably converge in a class-wise manner as adaptation proceeds.

\begin{table*}[h!]
\begin{center}
\footnotesize
\scalebox{0.9}{
\begin{tabular}{c|ccccccccccccccc|c}
\toprule
Method  & Gaus. & Shot & Impu. & Defo. & Glas. & Moti. & Zoom & Snow & Fros. & Fog & Brig. & Cont. & Elas. & Pixe. & Jpeg. & Average \\
\drule
Source & 43.31 &  34.34 &  43.78 &  8.32 &  52.34 &  16.72 &  8.12 &  19.31 &  20.07 &  13.02 &  \textbf{5.88} &  7.45 &  13.04 &  26.45 &  17.12 & 21.95\\
BN & 15.97 &  13.14 &  22.03 &  8.30 &  25.85 &  12.42 &  7.23 &  16.99 &  11.70 &  13.07 &  7.49 &  7.49 &  13.57 &  9.25 &  12.38 &  13.13\\
PL & 15.81 &  13.03 &  21.66 &  8.21 &  25.62 &  12.23 &  7.18 &  16.84 &  11.64 &  12.88 &  7.55 &  7.39 &  13.59 &  9.27 &  12.28 &  13.01 \\
FR-Online$^{\dagger}$ & 15.93 & 13.15 & 21.87 & 8.27 & 25.61 & 12.38 & 7.26 & 16.98 & 11.72 & 13.05 & 7.52 & 7.50 & 13.56 & 9.26 & 12.37 & 13.10   \\
TFA-Online$^{\dagger}$ & 14.69 & 12.29 & 18.03 & 7.57 & 22.84 & 11.19 & 6.59 & 14.77 & 10.48 & 10.62 & 6.68 & 6.89 & 11.90 & 9.08 & 11.03 & 11.64   \\
TTT++-Online$^{\dagger}$ & 15.20 & 12.57 & 17.33 & 7.61 & 23.07 & 10.72 & 6.62 & 13.31 & 10.63 & 9.87 & 6.14 & \textbf{6.29} & 12.13 & 8.95 & 11.92 & 11.49  \\
EATA$^\dagger$ & 15.52 & 12.73 & 21.08 & 8.16 & 24.72 & 12.07 & 7.12 & 16.50 & 11.54 & 12.62 & 7.42 & 7.41 & 13.34 & 9.07 & 12.04 & 12.76 \\
TENT$^\dagger$ & 14.51 &  11.98 &  19.12 &  7.69 &  22.81 &  10.96 &  7.10 &  15.26 &  11.18 &  11.03 &  7.05 &  7.03 &  12.69 &  8.75 &  11.64 & 11.92\\
\midrule
\textbf{CAFA (Ours)} & \textbf{12.73} &  \textbf{10.51} &  \textbf{16.71} &  \textbf{6.66} &  \textbf{20.34} &  \textbf{9.73} &  \textbf{6.07} &  \textbf{12.82} &  \textbf{9.51} &  \textbf{8.98} &  6.14 &  6.35 &  \textbf{11.25} & \textbf{7.76} &  \textbf{10.31} & \textbf{10.39} \\
\midrule
Source &  76.96 &  69.67 &  79.91 &  32.50 &  82.34 &  46.32 &  32.75 &  50.96 &  51.77 &  44.18 &  \textbf{26.44} &  32.44 &  41.08 &  52.03 &  46.34 & 51.05 \\
BN &  44.93 &  40.95 &  52.39 &  31.68 &  57.40 &  37.53 &  30.28 &  45.78 &  38.16 &  41.63 &  30.30 &  31.13 &  40.37 &  32.93 &  37.22 & 39.51\\
PL & 44.54 &  40.42 &  51.50 &  31.63 &  56.51 &  37.05 &  29.95 &  45.38 &  37.78 &  40.93 &  30.16 &  30.82 &  40.04 &  32.56 &  36.85 & 39.07\\
FR-Online$^{\dagger}$ & 44.92 & 40.99 & 52.39 & 31.72 & 57.38 & 37.55 & 30.33 & 42.01 & 38.14 & 41.62 & 30.32 & 31.17 & 40.36 & 32.90 & 37.19 & 39.27  \\
TFA-Online$^{\dagger}$& 42.39 & 37.82 & 48.26 & 30.03 & 54.24 & 34.70 & 28.41 & 41.59 & 35.60 & 36.74 & 28.27 & 29.78 & 36.50 & 31.93 & 35.62 & 36.79 \\
TTT++-Online$^{\dagger}$ & 41.16 & 37.15 & 47.38  & 28.74 & 53.18 & 32.81 & 27.95 & 40.65 & 34.85 & 34.30 & 27.34 & 27.83 & 35.41 & 30.59 & 34.99 & 35.62  \\
EATA$^{\dagger}$ & 42.74 & 38.76 & 48.48 & 30.56 & 53.87 & 35.45 & 29.31 & 43.17 & 36.66 & 38.38 & 29.30 & 30.04 & 38.25 & 31.75 & 35.72 & 37.50 \\
TENT$^{\dagger}$ &  41.28 &  37.10 &  46.18 &  29.06 &  51.82 &  33.62 &  28.08 &  41.18 &  35.33 &  35.01 &  27.68 &  28.48 &  36.12 &  30.63 &  34.40 & 35.73\\
\midrule
\textbf{CAFA (Ours)}& \textbf{39.11} &  \textbf{35.71} &  \textbf{44.17} &  \textbf{27.83} &  \textbf{49.22} &  \textbf{32.16} &  \textbf{27.56} &  \textbf{39.36} &  \textbf{34.05} &  \textbf{33.22} &  26.61 &  \textbf{27.49} &  \textbf{34.97} & \textbf{29.68} & \textbf{33.47} & \textbf{34.31}\\
\bottomrule
\end{tabular}}
\vspace{0.1cm}
\caption{Classification error (\%) on the CIFAR10-C (upper group) and CIFAR100-C (lower group) datasets with severity level 4 corruptions.
$^\dagger$ denotes the results obtained from the official codes.}
\vspace{-0.2cm}
\label{suptab:cifar_lv4}
\end{center}
\end{table*}


\begin{table*}[h!]
\begin{center}
\footnotesize
\scalebox{0.9}{
\begin{tabular}{c|ccccccccccccccc|c}
\toprule
Method  & Gaus. & Shot & Impu. & Defo. & Glas. & Moti. & Zoom & Snow & Fros. & Fog & Brig. & Cont. & Elas. & Pixe. & Jpeg. & Average \\
\drule
Source & 36.96 &  28.00 &  26.86 &  \textbf{5.73} &  35.53 &  16.68 &  7.54 &  16.89 &  18.54 &  8.98 &  \textbf{5.64} &  6.47 &  \textbf{7.69} &  13.10 &  15.54 & 16.68 \\
BN & 13.75 &  11.97 &  16.01 &  7.39 &  16.74 &  12.36 &  7.33 &  14.89 &  11.63 &  10.41 &  7.03 &  7.23 &  9.50 &  8.43 &  11.51 & 11.08 \\
PL &   13.59 &  11.90 &  15.81 &  7.35 &  16.58 &  12.23 &  7.44 &  14.75 &  11.41 &  10.36 &  6.97 &  7.13 &  9.41 &  8.32 &  11.48 & 10.98\\
FR-Online$^{\dagger}$ & 13.70 & 11.98 & 16.01 & 7.40 & 16.73 & 12.31 & 7.38 & 14.88 & 11.61 & 10.40 & 7.04 & 7.26 & 9.45 & 8.43 & 11.47 & 11.07  \\
TFA-Online$^{\dagger}$ & 12.71 & 11.00 & 13.56 & 6.65 & 14.96 & 10.96 & 6.72 & 12.76 & 10.30 & 8.77 & 6.35 & 6.49 & 8.33 & 7.81 & 10.42 & 9.85 \\
TTT++-Online$^{\dagger}$ & 13.41 & 11.16 & 12.71 & 6.18 & 15.14 & 10.55 & 6.45 & 11.87 & 10.32 & 7.78 & 5.80 & \textbf{6.11} & 8.38 & 7.36 & 10.80 & 9.60 \\
EATA$^{\dagger}$ & 13.32 & 11.79 & 15.51 & 7.35 & 16.41 & 12.07 & 7.19 & 14.47 & 11.37 & 10.09 & 6.96 & 7.10 & 9.30 & 8.34 & 11.22 & 10.83 \\
TENT$^\dagger$ & 12.47 &  11.29 &  14.12 &  6.86 &  15.47 &  11.26 &  6.95 &  13.40 &  10.81 &  9.09 &  6.52 &  6.87 &  8.99 &  7.93 &  10.84 & 10.19 \\
\midrule
\textbf{CAFA (Ours)} & \textbf{10.90} &  \textbf{10.01} &  \textbf{11.89} & 5.95 &  \textbf{13.23} & \textbf{9.77} &  \textbf{5.99} &  \textbf{11.38} &  \textbf{9.47} &  \textbf{7.54} &  5.93 &  6.28 &  8.00 &  \textbf{7.19} &  \textbf{9.60} & \textbf{8.88} \\
\midrule
Source & 71.80 &  63.00 &  65.65 &  \textbf{26.15} &  73.00 &  46.43 &  31.03 &  48.51 &  48.84 &  35.35 &  \textbf{25.45} &  28.77 &  30.76 &  38.34 &  43.74 & 45.12\\
BN &  41.70 &  38.35 &  43.93 &  29.59 &  44.93 &  37.70 &  29.89 &  42.60 &  37.45 &  37.00 &  29.72 &  30.52 &  34.24 &  31.80 &  35.40 & 36.32 \\
PL & 41.26 &  38.08 &  43.45 &  29.59 &  44.68 &  37.35 &  29.69 &  42.33 &  37.33 &  36.50 &  29.55 &  30.21 &  33.80 &  31.58 &  35.11 & 36.03\\
FR-Online$^{\dagger}$ & 41.70 & 38.36 & 43.94 & 29.57 & 44.88 & 37.67 & 29.91 & 42.62 & 37.46 & 37.04 & 29.72 & 30.50 & 34.23 & 31.81 & 35.42 & 36.32 \\
TFA-Online$^{\dagger}$ & 39.27 & 35.77 & 40.52 & 27.98 & 42.05 & 34.99 & 28.51 & 38.72 & 35.01 & 32.90 & 27.83 & 28.81 & 31.43 & 30.60 & 34.38 & 33.92 \\
TTT++-Online$^{\dagger}$ & 38.39 & 34.51 & 38.95 & 26.52 & 41.53 & 33.65 & 27.64 & 37.75 & 34.06 & 31.32 & 26.78 & 27.22 & \textbf{30.18} & 29.46 & 33.26 & 32.75 \\
EATA$^{\dagger}$ & 40.00 & 36.57 & 40.94 & 28.82 & 42.28 & 35.85 & 28.96 & 40.43 & 36.06 & 34.53 & 29.06 & 29.44 & 32.90 & 30.56 & 34.25 & 34.71 \\
TENT$^{\dagger}$ &  38.37 &  34.72 &  39.50 &  27.41 &  40.91 &  33.81 &  27.99 &  39.10 &  34.69 &  31.97 &  27.36 &  27.94 &  31.80 &  29.09 &  32.94 & 33.17 \\
\midrule
\textbf{CAFA (Ours)}& \textbf{36.97} &  \textbf{33.53} &  \textbf{37.41} &  26.21 &  \textbf{38.96} &  \textbf{32.46} &  \textbf{27.26} &  \textbf{36.82} &  \textbf{33.64} &  \textbf{30.13} &  26.53 &  \textbf{27.07} &  30.20 &  \textbf{28.23} &  \textbf{32.05} & \textbf{31.83} \\
\bottomrule
\end{tabular}}
\vspace{0.1cm}
\caption{Classification error (\%) on the CIFAR10-C (upper group) and CIFAR100-C (lower group) datasets with severity level 3 corruptions.
$^\dagger$ denotes the results obtained from the official codes.}
\vspace{-0.2cm}
\label{suptab:cifar_lv3}
\end{center}
\end{table*}


\begin{table*}[h!]
\begin{center}
\footnotesize
\scalebox{0.9}{
\begin{tabular}{c|ccccccccccccccc|c}
\toprule
Method  & Gaus. & Shot & Impu. & Defo. & Glas. & Moti. & Zoom & Snow & Fros. & Fog & Brig. & Cont. & Elas. & Pixe. & Jpeg. & Average \\
\drule
Source & 24.43 &  14.91 &  18.87 &  \textbf{5.32} &  37.16 &  11.58 &  6.75 &  18.33 &  11.29 &  6.76 &  \textbf{5.38} &  5.94 &  \textbf{7.06} &  10.35 &  14.42 & 13.24\\
BN & 11.24 &  8.92 &  12.66 &  7.13 &  16.28 &  10.17 &  7.08 &  12.77 &  9.45 &  8.43 &  7.02 &  7.14 &  9.50 &  8.35 &  10.71 & 9.79 \\
PL &  11.17 &  8.81 &  12.71 &  6.92 &  16.27 &  10.10 &  7.15 &  12.67 &  9.43 &  8.41 &  7.02 &  6.96 &  9.45 &  8.31 &  10.66 & 9.74\\
FR-Online$^{\dagger}$ & 11.21 & 8.91 & 12.61 & 7.11 & 16.26 & 10.18 & 7.10 & 12.77 & 9.46 & 8.41 & 7.04 & 7.14 & 9.49 & 8.35 & 10.66 & 9.78  \\
TFA-Online$^{\dagger}$ & 10.29  & 8.09 & 10.85 & 6.35 & 14.87 & 8.97 & 6.43 & 11.29 & 8.27 & 7.47 & 6.21 & 6.40 & 8.39 & 7.42 & 9.61 & 8.73 \\
TTT++-Online$^{\dagger}$ & 10.47 & 7.96 & 10.16 & 5.89 & 14.92 & 8.63 & 6.26 & 10.40 & 8.26 & 6.80 & 5.78 & \textbf{5.86} & 7.84 & 7.07 & 9.75 & 8.40 \\
EATA$^\dagger$ & 11.12 & 8.83 & 12.36 & 7.00 & 15.86 & 9.83 & 7.01 & 12.52 & 9.29 & 8.30 & 6.91 & 7.01 & 9.25 & 8.27 & 10.38 & 9.60 \\
TENT$^\dagger$ & 10.52 &  8.15 &  11.45 &  6.65 &  15.03 &  9.64 &  6.67 &  11.18 &  8.69 &  7.44 &  6.50 &  6.72 &  8.80 &  7.82 &  9.85 & 9.01  \\
\midrule
\textbf{CAFA (Ours)} &  \textbf{9.22} &  \textbf{7.51} &  \textbf{9.65} &  5.78 &  \textbf{13.22} &  \textbf{8.20} &  \textbf{5.86} &  \textbf{9.71} &  \textbf{7.57} &  \textbf{6.55} &  5.87 &  6.02 &  7.64 &  \textbf{6.95} &  \textbf{8.90} & \textbf{7.91} \\
\midrule
Source & 59.17 &  44.40 &  54.99 &  \textbf{24.91} &  73.55 &  37.45 &  28.70 &  49.25 &  37.83 &  29.75 &  \textbf{25.01} &  26.84 &  \textbf{29.35} &  34.25 &  41.17 & 39.77\\
BN &  37.60 &  33.23 &  39.55 &  29.11 &  44.32 &  33.91 &  29.63 &  38.89 &  33.89 &  33.74 &  29.05 &  29.74 &  33.21 &  30.94 &  34.03 & 34.06\\
PL &  37.40 &  32.92 &  39.01 &  28.88 &  43.91 &  33.56 &  29.45 &  38.67 &  33.68 &  33.42 &  28.86 &  29.48 &  32.90 &  30.71 &  33.82 & 33.78 \\
FR-Online$^{\dagger}$ & 37.61 & 33.23 & 39.56 & 29.11 & 44.34 & 33.91 & 29.62 & 38.89 & 33.94 & 33.69 & 29.05 & 29.75 & 33.21 & 30.91 & 34.05 & 34.06  \\
TFA-Online$^{\dagger}$ & 35.27 & 31.10 & 36.71 & 27.86 & 42.04 & 32.00 & 28.65 & 36.03 & 31.18 & 30.22 & 27.60 & 28.32 & 30.65 & 29.30 & 33.11 & 32.00 \\
TTT++-Online$^{\dagger}$ & 33.60 & 29.74 & 34.88 & 25.90 & 41.21 & 30.64 & 26.97 & 34.86 & 30.17 & 28.89 & 26.19 & 26.68 & 29.57 & 28.29 & 31.84 & 30.63 \\
EATA$^{\dagger}$ & 36.01 & 32.23 & 37.20 & 28.17 & 41.81 & 32.72 & 28.73 & 37.01 & 32.60 & 32.17 & 28.13 & 28.88 & 31.80 & 29.69 & 32.92 & 32.67 \\
TENT$^{\dagger}$ &  34.29 &  30.72 &  35.37 &  27.22 &  40.35 &  31.03 &  27.54 &  35.55 &  31.05 &  29.80 &  26.84 &  27.60 &  30.54 &  28.43 &  31.82 & 31.21\\
\midrule
\textbf{CAFA (Ours)}& \textbf{33.02} &  \textbf{29.38} &  \textbf{33.75} &  26.09 &  \textbf{38.34} &  \textbf{29.63} &  \textbf{26.73} &  \textbf{33.47} &  \textbf{29.79} &  \textbf{28.18} &  25.94 &  \textbf{26.45} &  29.58 &  \textbf{27.64} &  \textbf{30.95} & \textbf{29.93} \\
\bottomrule
\end{tabular}}
\vspace{0.1cm}
\caption{Classification error (\%) on the CIFAR10-C (upper group) and CIFAR100-C (lower group) datasets with severity level 2 corruptions.
$^\dagger$ denotes the results obtained from the official codes.}
\vspace{0.1cm}
\label{suptab:cifar_lv2}
\end{center}
\end{table*}


\begin{table*}[h!]
\begin{center}
\footnotesize
\scalebox{0.9}{
\begin{tabular}{c|ccccccccccccccc|c}
\toprule
Method  & Gaus. & Shot & Impu. & Defo. & Glas. & Moti. & Zoom & Snow & Fros. & Fog & Brig. & Cont. & Elas. & Pixe. & Jpeg. & Average \\
\drule
Source & 14.00 &  10.09 &  11.62 &  \textbf{5.24} &  38.69 &  7.75 &  7.40 &  9.48 &  8.14 &  \textbf{5.74} &  \textbf{5.37} &  \textbf{5.47} &  \textbf{7.69} &  6.69 &  10.51 & 10.26\\
BN &  8.89 &  8.08 &  10.17 &  7.04 &  16.16 &  8.55 &  7.55 &  9.44 &  7.66 &  7.47 &  7.00 &  6.92 &  9.81 &  7.50 &  8.76 & 8.73\\
PL &  8.83 &  7.90 &  10.17 &  6.94 &  16.05 &  8.42 &  7.59 &  9.40 &  7.54 &  7.41 &  6.93 &  6.81 &  9.77 &  7.38 &  8.70 & 8.66\\
FR-Online$^{\dagger}$ & 8.92 & 8.08 & 10.19 & 7.03 & 16.12 & 8.57 & 7.53 & 9.48 & 7.67 & 7.49 & 7.00 & 6.92 & 9.80 & 7.47 & 8.77 & 8.74  \\
TFA-Online$^{\dagger}$ & 7.94 & 7.24 & 8.95 & 6.26 & 14.38 & 7.63 & 7.11 & 8.37 & 7.10 & 6.45 & 6.23 & 6.02 & 8.64 & 6.75 & 7.69 & 7.79 \\
TTT++-Online$^{\dagger}$ & 8.23 & 6.97 & 8.52 & 5.61 & 15.17 & 7.43 & 6.84 & \textbf{7.66} & 6.55 & 5.95 & 5.61 & 5.80 & 8.39 & 6.65 & 7.86 & 7.55 \\
EATA$^\dagger$ & 8.81 & 7.86 & 10.12 & 6.99 & 15.70 & 8.43 & 7.54 & 9.31 & 7.52 & 7.40 & 6.91 & 6.81 & 9.60 & 7.38 & 8.61 & 8.60 \\
TENT$^\dagger$ & 8.25 &  7.50 &  9.71 &  6.54 &  14.75 &  7.94 &  7.15 &  8.71 &  6.94 &  6.62 &  6.43 &  6.48 &  9.35 &  7.29 &  8.37 & 8.14\\
\midrule
\textbf{CAFA (Ours)} & \textbf{7.37} &  \textbf{6.54} &  \textbf{8.36} &  5.80 &  \textbf{13.07} &  \textbf{6.99} &  \textbf{6.34} &  7.84 &  \textbf{6.38} &  6.04 &  5.70 &  5.76 &  8.03 &  \textbf{6.35} &  \textbf{7.44} & \textbf{7.20} \\
\midrule
Source & 42.81 &  35.37 &  40.39 &  \textbf{24.70} &  74.15 &  31.29 &  29.80 &  35.16 &  31.07 &  \textbf{25.20} &  \textbf{24.79} &  \textbf{25.23} &  \textbf{30.04} &  28.64 &  34.83 & 34.23\\
BN & 32.74 &  30.64 &  34.21 &  28.69 &  44.57 &  32.11 &  30.34 &  33.10 &  30.59 &  29.69 &  28.55 &  28.90 &  34.23 &  29.94 &  31.52 & 31.99\\
PL &  32.61 &  30.50 &  33.70 &  28.52 &  44.10 &  31.83 &  30.19 &  32.82 &  30.52 &  29.25 &  28.46 &  28.61 &  33.84 &  29.72 &  31.27 & 31.73\\
FR-Online$^{\dagger}$ & 32.74 & 30.67 & 34.22 & 28.71 & 44.58 & 32.15 & 30.35 & 33.08 & 30.62 & 29.68 & 28.58 & 28.92 & 34.21 & 29.95 & 31.53 & 32.00  \\
TFA-Online$^{\dagger}$ & 30.64 & 28.78 & 31.88 & 27.43 & 41.80 & 30.58 & 29.04 & 30.70 & 29.27 & 27.84 & 27.31 & 27.56 & 31.11 & 28.58 & 30.29  & 30.19 \\
TTT++-Online$^{\dagger}$ & 29.63 & 28.25 & 30.33 & 25.63 & 41.00 & 29.07 & 27.36 & 29.76 & 27.81 & 26.35 & 26.17 & 26.14 & 30.11 & 27.34 & 29.15 & 28.94 \\
EATA$^{\dagger}$ & 31.73 & 29.93 & 32.78 & 27.82 & 42.00 & 31.03 & 29.45 & 31.99 & 29.76 & 28.40 & 27.89 & 28.01 & 32.85 & 29.01 & 30.53 & 30.88 \\
TENT$^{\dagger}$ &  30.07 &  28.53 &  31.01 &  26.84 &  40.33 &  29.45 &  28.18 &  30.56 &  28.49 &  27.14 &  26.71 &  27.07 &  31.33 &  27.76 &  29.12 & 29.51\\
\midrule
\textbf{CAFA (Ours)}& \textbf{28.82} &  \textbf{27.70} &  \textbf{30.09} &  25.84 &  \textbf{38.38} &  \textbf{28.02} &  \textbf{27.09} &  \textbf{29.06} &  \textbf{27.36} &  26.23 &  25.80 &  25.91 &  30.33 &  \textbf{26.94} &  \textbf{28.63} & \textbf{28.41}\\
\bottomrule
\end{tabular}}
\end{center}
\vspace{-0.3cm}
\caption{Classification error (\%) on the CIFAR10-C (upper group) and CIFAR100-C (lower group) datasets with severity level 1 corruptions.
$^\dagger$ denotes the results obtained from the official codes.}
\vspace{-1.0cm}
\label{suptab:cifar_lv1}
\end{table*}


\begin{table*}[h!]
\begin{center}
\footnotesize
\scalebox{0.9}{
\begin{tabular}{c|ccccccccccccccc|c}
\toprule
Trial & Gaus. & Shot & Impu. & Defo. & Glas. & Moti. & Zoom & Snow & Fros. & Fog & Brig. & Cont. & Elas. & Pixe. & Jpeg. & Average \\
\drule
0 & 14.05 &  13.06 &  21.35 &  8.07 &  20.45 &  11.10 &  6.96 &  11.92 &  11.31 &  13.22 &  7.06 &  7.14 &  16.05 &  9.65 &  11.70 & 12.21\\
1 & 13.79 &  12.90 &  21.05 &  8.13 &  21.08 &  10.85 &  6.92 &  11.92 &  11.20 &  13.10 &  6.83 &  7.22 &  16.03 &  9.63 &  11.57 & 12.15\\
2 & 14.07 &  12.60 &  21.01 &  7.89 &  20.73 &  10.71 &  6.86 &  11.87 &  11.24 &  13.46 &  6.92 &  6.94 &  16.16 &  9.80 &  11.55 & 12.12\\
3 & 13.80 &  12.99 &  20.83 &  8.00 &  20.62 &  10.55 &  6.71 &  11.89 &  11.64 &  13.68 &  7.11 &  7.04 &  16.22 &  9.82 &  11.25 & 12.14\\
4 & 13.70 &  12.71 &  21.07 &  8.09 &  20.94 &  10.68 &  6.87 &  11.96 &  11.41 &  13.82 &  7.04 &  7.05 &  15.90 &  9.67 &  11.68 & 12.17\\
\midrule
std & 0.17 & 0.19 &	0.19 & 0.09 &	0.25&	0.21&	0.10&	0.03&	0.18&	0.30&	0.11&	0.11&	0.12&	0.09&	0.18&	0.03 \\
\bottomrule
\end{tabular}}
\vspace{0.1cm}
\caption{Classification error (\%) on the severity level 5 corruptions in the CIFAR10-C dataset with different random seeds.}
\vspace{-0.6cm}
\label{suptab:cifar_randomseeds}
\end{center}
\end{table*}

\clearpage

\clearpage

{\small
\bibliographystyle{ieee_fullname}
\bibliography{egbib}
}

\end{document}